\pdfoutput=1

\documentclass[journal]{IEEEtran}
\usepackage{ifpdf}
\usepackage{cite}
   \usepackage[pdftex]{graphicx}
\graphicspath{{figures/}}
\DeclareGraphicsExtensions{.pdf,.jpeg,.png,.jpg}
\usepackage{amsmath}
\usepackage{algorithm,algorithmic}

  \usepackage[caption=false,font=footnotesize]{subfig}
\usepackage{url}
\usepackage[capitalise]{cleveref}
\usepackage{tabularx}
\usepackage{booktabs}
\usepackage{framed}
\usepackage{pdfpages}
\usepackage[percent]{overpic}

\usepackage{xcolor}

\newcommand{\revisionOne}[1]{#1}
\newcommand{\revisionOnePar}[1]{#1}

\makeatletter
\newcommand\thefontsize[1]{{#1 The current font size is: \f@size pt\par}}
\makeatother

\begin{document}
\title{Genetic Programming for Evolving a Front of Interpretable Models for Data Visualisation}

\author{Andrew~Lensen,~\IEEEmembership{Member,~IEEE,}
	Bing~Xue,~\IEEEmembership{Member,~IEEE,}
	and~Mengjie~Zhang,~\IEEEmembership{Fellow,~IEEE}% <-this % stops a space
	\thanks{This work was supported in part by the Marsden Fund of New Zealand Government under Contracts VUW1509 and VUW1615, the Science for Technological Innovation Challenge (SfTI) fund under grant E3603/2903,  and the University Research Fund at Victoria University of Wellington grant number 216378/3764 and 223805/3986. 
	
	The authors are with the Evolutionary Computation Research Group, Victoria University of Wellington, Wellington 6140, New Zealand (e-mail: andrew.lensen@ecs.vuw.ac.nz; bing.xue@ecs.vuw.ac.nz; mengjie.zhang@ecs.vuw.ac.nz).	\vspace{-1em}}      
}

\markboth{Journal of \LaTeX\ Class Files,~Vol.~14, No.~8, August~2015}%
{Shell \MakeLowercase{\textit{et al.}}: Bare Demo of IEEEtran.cls for IEEE Journals}



\maketitle

\begin{abstract}
	Data visualisation is a key tool in data mining for understanding big datasets. Many visualisation methods have been proposed, including the well-regarded state-of-the-art method t-Distributed Stochastic Neighbour Embedding. However, the most powerful visualisation methods have a significant limitation: the manner in which they create their visualisation from the original features of the dataset is completely opaque. Many domains require an understanding of the data in terms of the original features; there is hence a need for powerful visualisation methods which use understandable models. In this work, we propose a genetic programming approach named GP-tSNE for evolving interpretable mappings from a dataset to high-quality visualisations. A multi-objective approach is designed that produces a variety of visualisations in a single run which give different trade-offs between visual quality and model complexity. Testing against baseline methods on a variety of datasets shows the clear potential of GP-tSNE to allow deeper insight into data than that provided by existing visualisation methods. We further highlight the benefits of a multi-objective approach through an in-depth analysis of a candidate front, which shows how multiple models can be analysed jointly to give increased insight into a dataset.
\end{abstract}

\begin{IEEEkeywords}
Visualisation, t-Distributed Stochastic Neighbourhood Embedding, Genetic Programming, Feature Selection.	\vspace{-.5em}
\end{IEEEkeywords}

% \fi
\IEEEpeerreviewmaketitle

%align float pages to top
\makeatletter
\setlength{\@fptop}{0pt}
\setlength{\@fpbot}{0pt plus 1fil}
\makeatother

\section{Introduction}
\IEEEPARstart{V}{isualisation} is a fundamental task in data mining which aims to represent data in a human-understandable graphical manner \cite{fayyad1996data}. Visualisation is often touted as a useful tool for allowing practitioners to gain an understanding of the data being analysed, but state-of-the-art visualisation methods (e.g.\ t-Distributed Stochastic Neighbour Embedding (t-SNE) \cite{maatenTSNE}) tend to be black boxes that give no insight into how the visualisation represents the original features of the data. Other methods such as autoencoders \cite{hinton2006reducing}  and parametric t-SNE \cite{maaten2009learning} produce a parametric mapping from the original dataset to the two-dimensional visualisation, but these use complex deep neural networks which are opaque to humans and provide little ``intuitive'' understanding. Even the most recent advances such as Uniform Manifold Approximation and Projection (UMAP) \cite{2018arXivUMAP} acknowledge significant weaknesses with regards to interpretability. As McInnes et al.\ note (emphasis ours):\\

\begin{quote}
	%\small
	%don't ask.
		%\fontsize{9pt}{10.8pt}\selectfont
			\vspace{-.5em}
	\fontsize{8.99999pt}{10.8pt}\selectfont
	``\textbf{For a number of uses [\textit{sic}] cases the interpretability of the reduced dimension results is of critical importance. Similarly to most non-linear dimension reduction techniques (including t-SNE and Isomap), UMAP lacks the strong interpretability} of Principal Component Analysis (PCA) and related techniques such a [\textit{sic}] NonNegative Matrix Factorization (NMF). In particular \textbf{the dimensions of the UMAP embedding space have no specific meaning}, unlike PCA where the dimensions are the directions of greatest variance in the source data. Furthermore, since UMAP is based on the distance between observations rather than the source features, it does not have an equivalent of factor loadings that linear techniques such as PCA, or Factor Analysis can provide. If strong interpretability is critical we therefore recommend linear techniques such as PCA and NMF.'' \cite[p.~35]{2018arXivUMAP}\\
\end{quote}
	\vspace{-.5em}
While such linear techniques as PCA and NMF provide strong interpretability, they are inherently limited in the quality of their output dimensions by their use of linear weightings only. Tree-based genetic programming (GP) is often recognised for its ability to directly model functions by taking inputs as leaves and producing an output at the root \cite{poli2008field} and is often used for dimensionality reduction in the form of feature construction (FC) \cite{neshatian2012filter}. Significant progress has been made on producing interpretable GP trees, which are simple enough to be understood by a human expert by using techniques such as parsimony pressure \cite{poli2008field} and multi-objective optimisation \cite{bleuler2001multiobjective}, with multi-objective approaches showing particularly good results due to their ability to produce a range of solutions with different levels of complexity \cite{wagner2012parsimony,cano2017multi}. Despite being clearly suitable for dimensionality reduction and evolving interpretable trees, GP has never been used to produce interpretable models that create understandable visualisations. 

We recently investigated the use of GP to do manifold learning (GP-MaL) for dimensionality reduction \cite{lensen2019can}. We found that GP was able to significantly reduce the size of feature sets while still retaining high-dimensional structure by producing sophisticated constructed features which used a variety of non-linear and conditional transformations. Our findings also highlighted the considerable potential of GP-MaL for visualisation, but that the original fitness function needed substantial refinement and that the complexity of the evolved models would need to be significantly reduced.

%\subsection{Goals}
In this paper we aim to significantly expand our previous work by using a multi-objective GP approach for visualisation which is expected to address the above limitations. To our knowledge, this is the first multi-objective GP method proposed which treats visualisation quality and model complexity as competing objectives. This paper will:

\begin{itemize}
	\item Propose a holistic set of functions and terminals for creating powerful and interpretable models for visualisation;
	\item Design a multi-objective approach to allow for the evolution of a solution front containing visualisations representing different levels of quality and complexity;
	\item Compare the quality and interpretability of the evolved visualisations with those produced by state-of-the-art visualisation methods on a range of datasets; and
	\item Perform an in-depth analysis of the trade-off between visualisation quality and tree complexity to demonstrate the unique advantages of the proposed approach.
\end{itemize}

This work is expected to provide a new and innovative approach which addresses the little-studied issue of model interpretability in visualisation. Of particular novelty is the in-depth analysis that will be performed that allows significant insight which is unattainable with existing single-solution black box approaches.
\section{Background}
\subsection{Dimensionality Reduction}
Dimensionality reduction (DR), the process of reducing the number of features/attributes in a dataset to improve understanding and  performance \cite{liu2012feature}, has continued to grow in importance as datasets become increasingly big and deep neural networks become more and more un-interpretable. Common techniques to address this problem include feature selection (FS) and feature construction (FC), which reduce the feature space through removing unwanted features or creating fewer, more complex meta-features, respectively. Evolutionary Computation (EC) methods have been applied to these NP-hard problems with significant success \cite{xue2015survey,neshatian2012filter}. In particular, tree-based GP has proved to be a natural fit for FC problems due to its functional and interpretable nature; GP-based FC has been used in classification \cite{tran2016genetic}, image analysis \cite{alsahaf2017automatically}, clustering \cite{lensen2017GPGC} and other domains \cite{Ahmed2014Multiple,hart2017hybrid}. 

Manifold learning (also known as non-linear dimensionality reduction) \cite{lee2007nonlinear} can be regarded as an unsupervised FC approach, where the task is to build a set of features which represent the non-linear manifold present in the high-dimensional feature space. One way of approaching this task is to attempt to build a \textit{function}, which maps the high-dimensional space to the low-dimensional manifold; such an approach could provide an understandable mapping between the two. Until our recent work \cite{lensen2019can} there had been no investigation into whether GP could be used to evolve such a mapping. 

\subsection{Machine Learning for Performing Visualisation}
The simplest commonly-used machine-learning visualisation methods are FS and Principal Component Analysis (PCA) \cite{jolliffe2011pca}. By using FS to select only two (or perhaps three) features, one can visualise a dataset by plotting the feature values of each instance along the x- and y-axes. PCA uses a more sophisticated approach, where it creates a series of ``principal components'' which are linear combinations of the feature set, such that each successive component is orthogonal to all the preceding components. By plotting the first two created components, a visualisation is created which is optimal when performing only linear transformations.

However, linear transformations fail to give clear visualisations of any underlying manifold/structure in a dataset beyond simple, low-dimensional data. Unfortunately, creating optimal non-linear transformations is an NP-hard problem, with many machine learning methods proposed as a result. The earliest methods include techniques such as Isomap \cite{tenenbaum2000isomap} and Local Linear Embedding (LLE) \cite{roweis2000lle}, but t-{SNE} \cite{maatenTSNE} is generally regarded as the mainstream state-of-the-art method.

t-SNE is an improvement to the previously proposed SNE method \cite{hinton2002sne}, which introduced a more nuanced probabilistic mapping from the high to low-dimensional spaces based on the similarity of neighbours in the high dimensional space. t-SNE improved upon SNE by introducing a cost function that could be better optimised by gradient-based optimisers; and by using a heavy-tailed t-distribution in place of a Gaussian to address the tendency of points in the low-dimensional space to be pushed together at the cost of separation between points, a characteristic known as the \textit{crowding problem}. A range of improvements to t-{SNE} have since been proposed, including a more efficient tree-based nearest-neighbour search \cite{maaten2014accelerating}, and a parametric version\cite{maaten2009learning}, but the cost function, which is the key aspect in our work, has remained relatively unchanged.

\subsection{Multi-objective Optimisation}
\label{moo}
Multi-objective optimisation (MO) is a technique used when a problem intrinsically has two (or more) \textit{conflicting} objectives between which a trade-off must be made by any solution to the problem. The quality of a solution in this context is often relative to the objective function values of other solutions. For example, given two candidate solutions $y,z$ to a problem with \revisionOne{$k$ objectives to be minimised}, then the following test can be used to determine if $y$ is strictly better than (or \textit{dominates}) $z$:
%\vspace{-.5em}
\begin{equation}
\forall i: f_i(y) \leq f_i(z) \text{ and } \exists j: f_j(y) < f_j(z)
%\vspace{-.5em}
\end{equation}
where \revisionOne{$i, j \in \{1, 2, \dots, k\}$} for $k$ objectives (in this paper, $k=2$). \revisionOne{A solution $z$ for which this does not hold true for all other solutions $y$} (i.e.\ it is \textit{non-dominated}) is called a \textit{Pareto-optimal} solution. \revisionOne{The set of Pareto-optimal solutions is often called the \textit{Pareto set}. The \textit{Pareto front} is then the image of the Pareto set in the objective space. In practice, it is infeasible or very difficult to find the true Pareto front, and so MO algorithms hence search for the best \textit{approximation front}}.

EC methods are some of the most successful and widely used approaches to MO as their population-based structure naturally allows for multiple non-dominated solutions to be found in one run. Of the Evolutionary Multi-objective Optimisation (EMO) algorithms, the Multi-Objective Evolutionary Algorithm based on Decomposition (MOEA/D) \cite{zhang2007moead} has seen widespread use in recent years. MOEA/D has been shown to produce \revisionOne{approximation} fronts with better spread and convergence than previous EMO methods\revisionOne{\cite{li2009mo}}. Multi-objective genetic programming (MOGP) approaches have seen significant success in recent years \cite{bhowan2013evolving,cano2017multi}.

% is perhaps the most popular for two-objective problems and is still regarded as a very strong and simple method despite being proposed in 2002. NSGA-II been widely used for multi-objective genetic programming \cite{bhowan2013evolving,cano2017multi}.

%NSGA-II works by constructing a series of dominance rankings, whereby an individual's ranking is based on the number of other solutions that dominate it. Solutions on the Pareto front have a ranking of $0$, with other solutions having higher (worse) rankings respectively. These rankings are then used as the fitness value in the evolutionary process, with selection methods such as tournament selection using these rankings/fitnesses to select solutions for breeding. The final output of the EC process is a set of unique, non-dominated solutions (i.e.\ the final Pareto front).

\subsection{Related Work}
\subsubsection{GP for Manifold Learning and Visualisation}
The only existing work we are aware of which used GP for MaL is our previous GP-MaL approach \cite{lensen2019can}. GP-MaL was proposed to tackle the general MaL problem, i.e. the reduction of a $d$-dimension high-dimensional space to a $d'$-dimension lower-dimensional space, where $d' \ll d$. A (single-objective) fitness function was proposed that measured how well manifold structure was preserved by measuring the preservation of \revisionOne{$n^\mathrm{th}$}-neighbour orderings in the lower-dimensional space. GP-MaL was fundamentally designed for dimensionality reduction for data mining tasks (e.g.\ classification), and so $d' > 2$ was primarily used to retain as much structure of the data as possible. Some initial application of GP-MaL for visualisation was investigated with encouraging results, but it was found that a more specific fitness function would be needed to improve visualisation quality, and that tree size would need to be addressed for visualisations to be understandable. 

Multi-objective GP approaches have been proposed to improve the visualisation quality of feature construction in a supervised learning context. The MOG3P method \cite{icke2011multi} aimed to produce solutions with a trade-off between three objectives: classifiability, visual interpretability, and semantic interpretability. Unlike in this work, MOG3P focused on performing \textit{supervised learning}, using class labels to guide the learning process. The three objectives also do not appear to be strictly conflicting: on a naive classification dataset, it would be possible to achieve perfect classification performance with a simple hyperplane split, which could be represented by a simple GP tree which produces a visualisation with two very distinct classes. Even on more complex datasets, there is an inherent relationship between the visual interpretability of a dataset (i.e.\ how well classes are separated), and the classification performance: given well-separated classes, one would expect correct classification to result. A later approach focused on classifiability and visual interpretability only, but used several different measures for each of these objectives \cite{cano2017multi}. This work also is a supervised approach, and so tackles a significantly different problem to that of this paper. The use of GP for visualisation has also been applied to combinatorial optimisation problems, such as job shop scheduling \cite{nguyen2018visualising}. 

\subsubsection{Other Approaches}
A variety of non-EC approaches have been proposed for MaL, some of which produce models that could theoretically be interpreted to understand the manifold in terms of the original features. Parametric t-SNE \cite{maaten2009learning} is a variation of t-SNE that allows for re-use of learnt t-SNE representations on future samples. Parametric t-SNE constructs a mapping from the high- to low-dimensional space using restricted Boltzmann machines to construct a pre-trained feed-forward neural network model. The neural network used over 10,000 neurons on the largest dataset, heavily restricting the potential for interpretation of the network. Autoencoders \cite{hinton2006reducing} are another neural-network based approach, which attempt to compress the representation of the data into as narrow of a middle hidden layer as possible such that the original data can be re-created from the concise representation. To do so, autoencoders use a number of layers of varying sizes to encode and decode the data. This provides a mapping to and from the learnt representation, but is unrealistic to interpret given the number of nodes and fully-connected topology. Attempts have been made to improve the interpretability of autoencoders \cite{sun2018evolving}, but success is fundamentally limited by the need for architectures which are differentiable. Self-organising maps (SOMs) \cite{kohonen1982som}, a variation of an unsupervised neural network, have been used for visualisation, but the number of weights scales with dimensionality and so their interpretability is limited on non-trivial datasets. The visualisation literature \cite{nonato2019multidimensional} discusses a few examples of mapping ``synthesised dimensions'' to original dimensions (in particular, page 4 of the supplementary material). These examples generally use a \textit{post hoc} approach, by varying the model's parameters or the instances \cite{faust2019dimreader} and analysing the effect, or using tools such as heatmaps. There is a distinct lack of literature that directly evolves simple models that use minimal unique features.

% Other FC work. Parametric t-SNE (limitations), autoencoders (limitations). Other model-based/parametric methods? 
\section{The Proposed Method: GP-tSNE}
Generally when applying GP to a problem, there are two main decisions to be made: what terminal and functional nodes are suitable, and how to formulate a fitness function to solve the problem. In addition, there are often other improvements made to the standard GP evolutionary process to further tailor it to the problem. Each of these three considerations are discussed in the following subsections. {%\color{red}
While each of these components builds on established work, the use of them together to produce a range of interpretable models producing high-quality visualisations is a new and substantial advance in this field. \revisionOne{The full source code of GP-tSNE is available online\footnote{\color{blue}\url{https://github.com/AndLen/gptsne}}}.}

\subsection{GP Architecture}
In order to project a high-dimensional space to a two-dimensional space for visualisation, it is important that a range of powerful and varied functions are available to the evolutionary process in order to produce compact but representative models. As exactly two dimensions are required for visualisation, we use a multi-tree approach where each individual contains two trees and each tree produces one dimension (i.e.\ the x- or y- axis) of the visualisation. A multi-tree representation is chosen rather than a co-operative co-evolution approach as the two trees must be tightly coupled (i.e.\ highly dependant) in order to give high-quality visualisations; co-operative co-evolution approaches tend to non-deterministically pair solutions from different sub-populations which greatly decreases the ability for such coupling to occur.

\subsubsection{Function Set}
Table \ref{terminalAndFunctionSet} lists the nine functions and four terminals used in GP-tSNE. The four arithmetic functions are standard, with the exception of the $n+$ and $-$ function which are the only functions that can take the zero node as input. This allows a variable number of inputs to these functions, which allows the evolutionary process to easily perform mutation that effectively removes whole trees (hopefully introns\revisionOne{: ``useless'' sub-trees whose output do not affect the tree output}) in the pursuit of model simplification. The $\div$ function performs protected division: if the denominator is $0$, it returns $1$.

\begin{table}
	\vspace{-1em}
	\centering
		\caption{The function and terminal sets of the proposed method.}
		\vspace{-.5em}
	\label{terminalAndFunctionSet}
	\begin{tabularx}{.7\linewidth}{@{}Xcr@{}}
		\toprule
		Function & No.\ of Inputs & Description\\
		\midrule
		 \multicolumn{3}{c}{Arithmetic Functions} \\
		 \midrule
		 $n+$ & \hspace{-8px}1--5 & Flexible Addition\\
 		 $-$ & \hspace{-8px}1--2 & Std.\ Subtraction\\
		 $\times$ & 2 & Std.\ Multiplication\\
		 $\div$ & 2 & Protected Division\\
		 \midrule
		 \multicolumn{3}{c}{Non-Linear Functions}\\
		 \midrule
		 Sigmoid & 1 & $\frac{1}{1+e^{-x}}$\\
		 ReLU &1&$\max(0,x)$\\
		 \midrule
		 \multicolumn{3}{c}{Conditional Functions}\\
		 \midrule
		 Max & 2 & $\max(x,y)$\\
		 Min & 2 & $\min(x,y)$\\
		 If & 3 & if $(x< 0)$: y; else $z$\\
		 \midrule
		 \multicolumn{3}{c}{Terminal Nodes}\\
		 \midrule
		 F$_i$ & 0 & $i^{th}$ feature value\\
		 NF$_i$ & 0 & \begin{tabular}[c]{@{}r@{}}Mean of $i^{th}$ feature values\\from 3-nearest neighbours\end{tabular} \\
		 %PCF$_j$ & 0 & \begin{tabular}[c]{@{}r@{}}Feature with $j^{th}$ largest\\ abs. weight in the 1st PC\end{tabular} \\
		 Constant & 0 & From $U[-1,1]$\\
		 Zero$^{\ast}$ & 0 & The number $0$\\
		 \bottomrule
	\end{tabularx}
	\vspace{-1.5em}
\end{table}

The sigmoid and ReLU functions are unary operators that are included to allow easy transformation of (linear) inputs into a non-linear output space. These were chosen based on inspiration from auto-encoders and other neural network methods. The three conditional operators provide a different kind of non-linear transform which is quite unique to GP due to their non-differentiability, and they are expected to allow a single tree to exhibit varied behaviour depending on its inputs.

\subsubsection{Terminal Set}
The F$_i$ terminal is commonly used in feature-based GP to use a feature as an input to a GP tree. Each feature in a dataset is assigned a distinct terminal which returns the values of the feature for a given instance. While this provides a great deal of flexibility to the EC search process, it does make the search space large for high numbers of features ($m$). To remedy this, we use PCA to produce the first PC on a dataset, and then select the $j$ features from that PC which have the highest-magnitude weights. Features with higher-magnitude weights contribute more to a PC, and therefore choosing the $j$ highest-magnitude features is a simple form of unsupervised feature selection. Each of these $j$ features have an increased likelihood of being chosen from the terminal set, which helps the EC search process focus on a set of promising features, while still allowing it to select other features with a smaller likelihood. In this work, we set $j = \sqrt{m}$, and each of the $j$ PCA-selected features have an \textit{additional} $j$ likelihood of being selected from the terminal set. For example, if $m = 64$ features, then $j = 8$: each of the 8 features chosen by PCA has ($1+8 = 9\times$ the likelihood) of being chosen from the terminal set. $j = \sqrt{m}$ is chosen so that the number of selected features increases slowly with the total number of features. Note that while we use PCA to weight promising features more strongly, we do not use the actual transformed PCs in our terminal set, as this would make the trees very difficult to interpret.

The NF$_i$ terminal serves two purposes: it provides a lower-noise version of a given feature, making trees less sensitive to noisy instances; and it allows a tree to encode local structure into the low-dimensional space more effectively by considering an instance's neighbourhood. While individual features (F$_i$) are suitable for embedding global structure, they are not as effective for preserving local structure. The three nearest-neighbours for each instance are pre-computed \revisionOne{by considering the ordering of the other instances by Euclidean distance}. 

The constant terminal is often used in GP, and is used here with a range of $[-1,+1]$ to allow different parts of the tree to have different impacts on the final output. The zero node is included solely for the $n+$ and $-$ function, and is not used by any other functions as it is either destructive or has no effect. The constant and zero nodes are chosen from the terminal set with a likelihood of $\lceil\frac{m}{10}\rceil$ as they are less likely to be useful to a function compared to a feature-based node. A summary of the terminal nodes' weightings are shown in \cref{table:weightings}. Note that the likelihood calculation is for a single instance in the case of a feature-based terminal, e.g.\ there are $m$ different $F_i$ terminals, each with a base likelihood of 1. \revisionOne{All of the numerical values, including the ``5'' in the flexible addition, the use of 3-nearest neighbours, and the likelihood values, were empirically chosen by testing on a range of representative datasets.}

\begin{table}
		\vspace{-1em}
	\centering
	\caption{Calculation of likelihood (l.h.) for each terminal to be selected from the terminal set.}
			\vspace{-.5em}
	\label{table:weightings}
	\begin{tabularx}{.7\linewidth}{@{}Xcr@{}}
		\toprule
		Terminal & L.h.\ & L.h.\ for $m=100$\\
		\midrule
		``Normal'' F$_i$ & 1 & 1\\
		PCA-selected F$_i$ & $1+\sqrt{m}$ & 11\\
		NF$_i$ & 1 & 1\\
		Constant & $\lceil{m}\div{10}\rceil$ & 10\\
		Zero & $\lceil{m}\div{10}\rceil$ & 10\\
		\bottomrule
	\end{tabularx}
	\vspace{-1em}
\end{table}
\vspace{-.25em}
\subsection{Multi-Objective Approach}
There is an intrinsic relationship in machine learning between the potential performance of a model and the complexity required to attain that performance. For example, the simplest model to separate two classes would be a decision boundary that simply thresholds at a certain point in space, whereas for three linearly-separable classes, at least two thresholds would be required (for some real $\alpha$, $\beta$: \ $Class_A < \alpha < Class_B < \beta < Class_C$). The same is true in visualisation: the more granular (specific) a visualisation is, the more complex the function used to produce that visualisation must be. To recreate the high-dimensional structure of a complex dataset in two dimensions would require two very big and complex GP trees. Each node removed from a tree decreases the accuracy with which the tree can reproduce the high-dimensional probability distribution (in the case of t-SNE). As an analogy, consider evolving a very complex polynomial function with GP: the fewer components (nodes) in the evolved functions, the fewer inflection points available to approximate the function.

In this work, we employ a multi-objective approach to produce a set of solutions that allow for a trade-off between visualisation quality and model interpretability to be chosen. This is strictly a more difficult problem than optimising only visualisation quality (i.e.\ as in t-SNE) as a model must be found which maps the high-dimensional to the low-dimensional space. We choose to use MOEA/D \cite{zhang2007moead} due to the reasons discussed in \cref{moo}. The first objective uses t-SNE to measure the visualisation quality of a GP tree; the second objective uses tree size as a proxy measure for the interpretability of the tree. These will be discussed in turn.
\vspace{-.25em}
\subsection{Objective 1: Visualisation Quality}
t-SNE uses conditional probabilities to represent the similarity between instances in a given dimensional space. Given two instances in the high-dimensional space, $x_i$ and $x_j$, the conditional probability $p_{j|i}$ that $x_j$ would be chosen as a neighbour of $x_i$ is defined as \cite{maatenTSNE}: 
\begin{equation}
\vspace{-.125em}
p_{j|i} = \frac{\revisionOne{\exp}(-\Vert x_i - x_j \Vert^2 /2\sigma_i^2)}{\sum_{k \neq l} \revisionOne{\exp}(-\Vert x_k - x_l \Vert^2 /2\sigma_i^2)}
\vspace{-.125em}
\end{equation}
given a Gaussian with variance $\sigma_i$ centred at $x_i$. t-SNE employs a symmetric approach where joint probability distributions are used; $p_{ij}$ is computed as the symmetrised conditional probabilities which for $n$ instances is defined as:
\begin{equation}
\vspace{-.125em}
p_{ij} = \frac{p_{j|i} + p_{i|j}}{2n}
\vspace{-.125em}
\end{equation}
In order to remedy the crowding problem (see \cite{maatenTSNE} for further details), t-SNE uses a slightly different approach in the low-dimensional space which is based on a Student t-distribution. The joint probabilities of two instances, $y_i$ and $y_j$, in the low-dimensional space, called $q_{ij}$ is computed as:
\begin{equation}
\vspace{-.125em}
q_{ij} = \frac{(1 + \Vert y_i - y_j \Vert^2)^{-1}}{\sum_{k \neq l}(1 + \Vert y_k - y_l \Vert^2)^{-1}}
\vspace{-.125em}
\end{equation}
The cost function (C), which measures the extent to which the low-dimensional probability distribution does \textbf{not} match the high-dimensional probability distribution, is the difference between the two distributions as measured using the sum of the Kullback-Leibler (KL) divergences:
\begin{equation}
\vspace{-.125em}
\label{costFunction}
\text{Cost} = KL(P\Vert Q) = \sum_i \sum_j p_{ij} \log\frac{p_{ij}}{q_{ij}}
\vspace{-.125em}
\end{equation}
We use this cost function as the first objective, which should be \textbf{minimized}. The parameter $\sigma$ is computed based on a perplexity of $40$ using the approach outlined in \cite{maatenTSNE}. We choose this cost function as it is well-tested and well-regarded in the MaL and visualisation literature. %However, it was developed with the constraint that it needed to be differentiable, and so it is likely there may be a non-differentiable formulation that is more powerful that could be used by GP as a fitness function --- this will be explored in future work in order to constrain the scope of this paper.

\subsection{Objective 2: Model Complexity}
A common issue encountered in GP is the production of bloated trees, where a GP tree is significantly bigger than is necessary to achieve a given level of fitness. Traditionally, there is no evolutionary pressure to encourage compact trees, and so trees may contain unnecessarily complex sub-trees, or in the worst case introns While being computationally inefficient, bloated trees are also much harder for humans to interpret and understand.

Parsimony pressure, which treats minimisation of model size as a (minor) objective in the optimisation process, is the most frequently method for controlling \textit{bloat} in GP \cite{poli2014parsimony}. Weighted sum approaches, where a small component of overall fitness is based on tree size, has been often used for attempting to control bloat, but choosing the right weighting is difficult and generally must be set empirically. More recently, multi-objective approaches have been proposed for addressing bloat, whereby tree size is used as a secondary objective to be minimised \cite{wagner2012parsimony}. This approach allows a trade-off between fitness and model complexity to be found according to the \revisionOne{approximation} front produced by the GP process, while also producing a range of solutions of varying complexity in a single GP run which can be compared by the user for better insight into the problem being tackled.

We use a simple formula for complexity in this work which is based on the number of nodes in each of the two trees, $T_a,T_b$, in a GP individual $I$: 
\begin{equation}
\vspace{-.125em}
\label{complexityFunction}
\revisionOne{\mathrm{Complexity}}(I) = \sum_{T \in I} \sum_{N_i \in T} \begin{cases}
0,& \text{if } N_i = \text{ZeroNode}\\
1,& \text{otherwise}
\end{cases}
\vspace{-.125em}
\end{equation}
A given node $N_i$ is counted towards the complexity unless it is a ZeroNode; these are not counted as they exist only to allow flexibility in the arity of the addition function, and can be removed from the tree structure when interpreting it.

%Standard NSGA-II was found to often produce a new population containing a large number of the simplest individuals (i.e.\ tree size 1--2), which causes a collapse of the front early on in the EC process. By enforcing this uniqueness constraint, a much more diverse and well-spread front is formed. 
\subsection{Optimisation of Tree Constants}
To further increase the visualisation quality without introducing additional model complexity, we use Particle Swarm Optimisation (PSO) \cite{kennedy1995particle} to fine-tune the Ephemeral random constants (ERCs) in each individual in the final front at the end of the evolutionary process. Standard GP has no ability to fine-tune its numerical parameters efficiently (as it randomly searches the parameter space); by employing PSO we can do so. Each ERC (from both trees) is allocated a dimension in the PSO representation, with the value of a given dimension representing how much the value of the given ERC varies from its original value. We use a very small range of initial position values ($[-0.15,0.15]$) and low minimum and maximum velocities ($-0.05$ and $0.05$) to focus the PSO search on fine-tuning the ERCs in the tree. One of the 30 particles is initialised with all values of $0$ (i.e.\ the original ERC values) so that the PSO search is guaranteed not to produce inferior solutions. The PSO search is single-objective (as the tree structure is fixed), with the fitness function being the same cost function used as the first objective in the GP search (\cref{costFunction}). PSO is only used at the end of the GP evolutionary process both due to its computational cost and to prevent GP from falling into local minima that would result from fine-tuning during evolution. 

Other techniques such as the Covariance Matrix Adaptation Evolution Strategy (CMA-ES) \cite{cmaes2003} have also been used for numerical optimisation, but PSO has been shown to give considerably better results on ill-conditioned functions\footnote{An ill-conditioned function is one with a high condition number: its output is overly sensitive to changes in the values of its variables.} \cite{hansen2011impacts}. For GP trees of sufficient complexity, it is expected they have the characteristics of an ill-conditioned function as a change in a given ERC value is likely to significantly change the output of the tree due to the interactions present across different sub-trees. Preliminary testing confirmed that PSO could optimise the random constants of the final GP individuals more effectively and consistently than CMA-ES.

\subsection{Other Considerations}
We use a slight variation to the standard MOEA/D, whereby we prevent the breeding process from producing individuals which have already been created in earlier generations\footnote{The breeding process tries up to 10 times to produce a non-duplicate individual --- otherwise it returns one of the parents, chosen randomly.}. While this added a small amount of overhead to the evolutionary process, it gave a much more efficient and effective learning process which was ultimately found to improve the diversity and quality of the final \revisionOne{approximated} front significantly. \revisionOne{The weight vectors in MOEA/D were scaled to encourage a uniformly distributed front. The cost objective was scaled to $[0,4]$ and complexity to $[0,4000]$.}

The evolutionary process was sped up through the use of multi-threaded evaluation, caching of quality values for previously-seen trees, and the use of linear algebra libraries which allow significantly faster matrix operations through native code libraries. \revisionOne{An overview of the GP-tSNE algorithm is shown in \cref{gptsneAlg}.
}
{
	%\vspace{-1em}
\begin{algorithm}[htb]
\small
			\revisionOnePar{
	\caption{\revisionOne{Overall GP-tSNE Algorithm}}
	
		\label{gptsneAlg}
		\begin{algorithmic}[1]
		\renewcommand{\algorithmicrequire}{\textbf{Input:}}
		\renewcommand{\algorithmicensure}{\textbf{Output:}}
		\REQUIRE Dataset: $X$, maximum generations: $G$
		\ENSURE  Approximated Pareto front $HOF$
		\STATE Randomly initialise population $P$
		\STATE $HOF \leftarrow \{\}$
		\\ \textit{Evolutionary Loop}:
		\FOR {$i = 1$ to $G$}
		
		\FOR {$j = 1$ to $|P|$}
		\STATE $A, B \leftarrow $ MOEAD$_{\mathrm{Selection}}(P_j)$ 
		\STATE $Offspring \leftarrow \{\}$
		\WHILE{$|Offspring| < 2$}
		\IF{$Rand() < CXPB$}
		\STATE $Child_A, Child_B = Crossover(A,B)$
		\ELSE
		\STATE $Child_A, Child_B = Mutate(A), Mutate(B)$
		\ENDIF
		\IF{$Unique(Child_B)$}
		\STATE $Offspring \leftarrow Offspring + Child_A$
		\ENDIF
		\IF{$Unique(Child_B)$}
		\STATE $Offspring \leftarrow Offspring + Child_B$
		\ENDIF
		\ENDWHILE
		\STATE $Vis_A = EvaluateInd(Offspring_A,X)$ 
		\STATE $Vis_B = EvaluateInd(Offspring_B,X)$
		\STATE Evaluate $CachedQuality(Vis)$ using \cref{costFunction}\\ and $Complexity(Offspring)$ using \cref{complexityFunction}
		\STATE $P \leftarrow $ MOEAD$_{\mathrm{UpdateFront}}(Offspring)$
		\ENDFOR
		\STATE $HOF \leftarrow NonDominated(HOF,P)$
		\ENDFOR
		\FOR{$j = 1$ to $|HOF|$}
		\STATE $HOF_j \leftarrow Optimise_{CMA-ES}(HOF_j)$
		\ENDFOR
		\RETURN $HOF$ 		
	\end{algorithmic} 
}
\end{algorithm}
\vspace{-1em}
}
\section{Experiment Setup}
The proposed GP-tSNE method was applied to a range of representative datasets which contain well-formed classes that lend well to visualisation. The nine datasets are summarised in Table \ref{table:datasets}, and have a range of numbers of instances, features, and classes. These datasets are from a number of different domains including general classification, biology, and image analysis. Most of these datasets were sourced from the UCI repository \cite{uci}. The standard t-{SNE} implementation \cite{maatenTSNE} was chosen as a baseline method as it has the same optimisation measure as GP-tSNE, and is regarded as the state-of-the-art in visualisation techniques. We used a standard perplexity value of $40$ (the same as in GP-tSNE). We also compare to our previous GP-MaL method \cite{lensen2019can} which was the first GP method to perform MaL, using a single-objective approach. GP-MaL parameter settings were unchanged from the paper.

		\renewcommand{\arraystretch}{1.1}
\begin{table}[t]
	\vspace{-1em}
	\caption{Classification datasets used for experiments.}
	\label{table:datasets}
	\centering
	\vspace{-.5em}
	\begin{tabularx}{.9\linewidth}{@{}Xrrr@{}}
		\toprule
		Dataset & Instances & Features & Classes\\
		\midrule
		Iris & 150 & 3 &3 \\
		Wine & 178 & 13 & 3 \\
		Dermatology & 358 & 34 & 6 \\
		Breast Cancer Wisconsin &  683 & 9 & 2 \\
		%Vehicle & 846 & 18 & 4 \\
		COIL20 & 1440 & 1024 & 20\\
		Isolet  & 1560 & 617 & 26 \\
		MFAT & 2000 & 649 & 10 \\
		MNIST 2-class & 2000 & 784 & 2\\
		Image Segmentation & 2310 & 19 & 7 \\ 
		
		\bottomrule
		
	\end{tabularx}
	\vspace{-1.5em}
\end{table}

We found that the use of a multi-objective approach necessitated a large number of generations (2,500) in order to allow the biggest trees (with the best performance) to be trained sufficiently. However, only a small population size of 100 individuals was needed to achieve a reasonable cover of the front, especially due to the technique used to breed unique individuals. The remaining parameters (Table \ref{table:parameterSettings}) are standard settings \revisionOne{--- tuning them further gave no change in performance.}

%; a maximum tree depth of 14 was found to be a good trade-off between allowing complex and powerful trees while not increasing the search space and computational cost unreasonably
 
Standard GP mutation is used, with one of the two trees in an individual being randomly chosen to be mutated. Crossover is performed by randomly selecting either the first or second tree to use, and then performing crossover between this tree in each parent --- this crossover occurs between the same axis of the visualisation.

	\begin{table}[t]
	%	\vspace{-1em}
	%	\captionsetup{position=top}
	%	\renewcommand{\arraystretch}{1.3}
	
	%	\footnotesize
	%\small
%	\vspace{-1em}
	\centering
	\caption{GP Parameter Settings.}
	\vspace{-.5em}
	\label{table:parameterSettings}
	%	\vspace{-.5em}
	\begin{tabularx}{0.95\linewidth}{ll X ll}
		
		\toprule
		Parameter& Setting && Parameter & Setting\\
		\cmidrule(r){1-2}  \cmidrule(l){4-5}
		Generations & 2500 && Population Size & 100\\
		Mutation & 20\% && Crossover & 80\% \\
		Min.\ Tree Depth & 2 && Max.\ Tree Depth & 14\\
		No.\ Trees & 2 && Pop. Initialisation & Half-and-half\\
		
		\bottomrule
	\end{tabularx}%
	%		\captionsetup{position=bottom}
	\vspace{-1em}
\end{table}

All three methods were run 30 times on each dataset. As our evaluation is primarily based on qualitative analysis (as is standard in visualisation \cite{maatenTSNE,2018arXivUMAP}), we chose the median result out of the 30 runs according to the objective function used by the method: t-{SNE}'s cost function for t-SNE and GP-tSNE, and the fitness function proposed in \cite{lensen2019can} for GP-MaL.

\revisionOnePar{t-SNE runs (by default) for up to 1,000 iterations, corresponding to 1,000 evaluations of the t-SNE cost function. GP-tSNE runs for 2500 generations with a population size of 100, giving 250,000 evaluations. Clearly, GP-tSNE is computationally more expensive, although the use of fitness caching and multi-threading reduces this expense. Also, GP-tSNE produces many visualisations in a single run, whereas t-SNE produces only one. On the biggest dataset, Image Segmentation, GP-tSNE takes 30 hours, whereas t-SNE takes around two minutes. We intend to significantly reduce this difference in future work, through the use of surrogate techniques and tuning of the GP design, but note that GP-tSNE is highly parallelisable, and so can run much quicker in practice.}

\section{Results and Discussion}
The results across each of the nine datasets are shown in Figs.\  \ref{fig:iris-median-PLOTS-PSO-PAPER}--\ref{fig:image-segmentation-median-PLOTS-PSO-PAPER}. Each figure contains eight plots, which we refer to as (a)--(h) reading left-to-right and top-to-bottom. Plot (a) shows the \revisionOne{attained approximation} front of the (median) GP result, with blue crosses showing each individual in the front, and the black line showing the shape of the front. The y and x-axes show each individual's value for Objective 1 (cost/visualisation quality) and 2 (model complexity) respectively. \revisionOne{The grey lines show the worst and best of the 30 GP-tSNE runs, which show the variance of the GP-tSNE algorithm. The green and yellow dotted lines represent the cost achieved by the baseline t-SNE and GP-MaL results respectively.} Plot (b) shows the same front, but with the x-axis truncated at 200 complexity, so as to focus on the distribution of the simpler/more interpretable trees. Plots (c)--(f) show representative visualisations --- coloured by class label --- produced by individuals at different levels of model complexity \revisionOne{(highlighted in red on the attained front)}: (c) is the simplest model \revisionOne{found by GP-tSNE} where each tree is a single feature (i.e.\ feature selection \revisionOne{(FS)}), (d) and (e) are two characteristic samples from along the front, and (f) is the most complex model, with the most accurate visualisation and lowest cost. \revisionOne{A small amount of jitter is added to visualisations with low complexity (less than 20) so that overlapping points can be better distinguished.} Plots (g) and (h) are the t-SNE and GP-MaL baseline results with median performance, where the \revisionOne{cost} written above the plots is calculated using t-SNE's objective function (the first objective of GP-tSNE), to allow for sensible comparisons.

It is only possible to show a small sample of the evolved visualisations in this paper, but one of the  significant advantages of GP-tSNE is its ability to produce a series of visualisations which progressively become more complex but more accurate representations of the high-dimensional feature space. To demonstrate this, we have included a video that shows each individual along the \revisionOne{approximation} front for each dataset, as the front is traversed from least to most complex. This is included with our submission, and is also available on YouTube\footnote{\vspace{-.5em}{\color{blue}\url{https://www.youtube.com/watch?v=K-z1jhBDHlY}}}. Specific visualisations can be studied in more detail by lowering the playback speed (e.g.\ to 25\% on YouTube).

\subsection{Specific Analysis}
\begin{figure}[]
		\vspace{-1em}
	\centering 
	\begin{overpic}[height=0.434\paperheight]{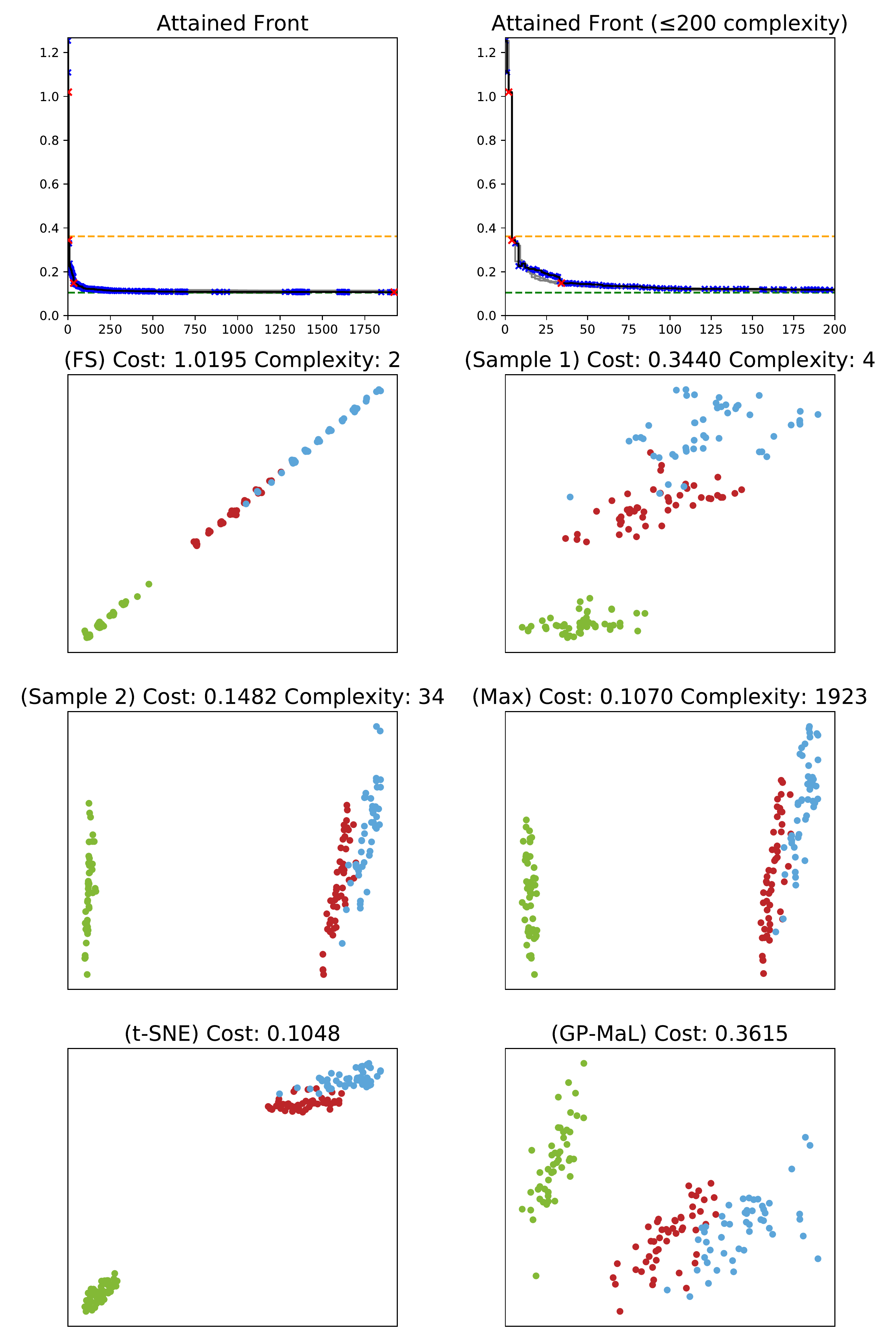}
		\put (-1,85.5) {(a)}
		\put (31.5,85.5) {(b)}
		\put (-1,60.5) {(c)}
		\put (31.5,60.5) {(d)}
		\put (-1,35.5) {(e)}
		\put (31.5,35.5) {(f)}
		\put (-1,10.5) {(g)}
		\put (31.5,10.5) {(h)}
	\end{overpic}
				\vspace{-1em}
	\caption{Iris.}
	\vspace{-1em}
	\label{fig:iris-median-PLOTS-PSO-PAPER}
\end{figure}

On the simplest dataset, Iris (\cref{fig:iris-median-PLOTS-PSO-PAPER}), all three methods provide a clear 2D visualisation, although t-SNE and GP-tSNE more distantly separate the green class. The Sample 1 visualisation for GP-tSNE is quite similar to that produced by GP-MaL when rotated, but with less continuous spacing of points, likely due to the trees being too small to convert the low-granularity feature space to a more complex output space. The front shows that diminishing returns are quickly found on this dataset: after {\raise.17ex\hbox{$\scriptstyle\sim$}}34 tree size (sample 2), quality only gradually improves as tree size increases. Indeed, the difference between the visualisations at complexity 34 and 1923 are very minor --- clearly GP-tSNE can capture the bulk of the structure in the simple Iris dataset using a simple model.
\begin{figure}[]
	\vspace{-1em}
	\centering
	\begin{overpic}[height=0.434\paperheight]{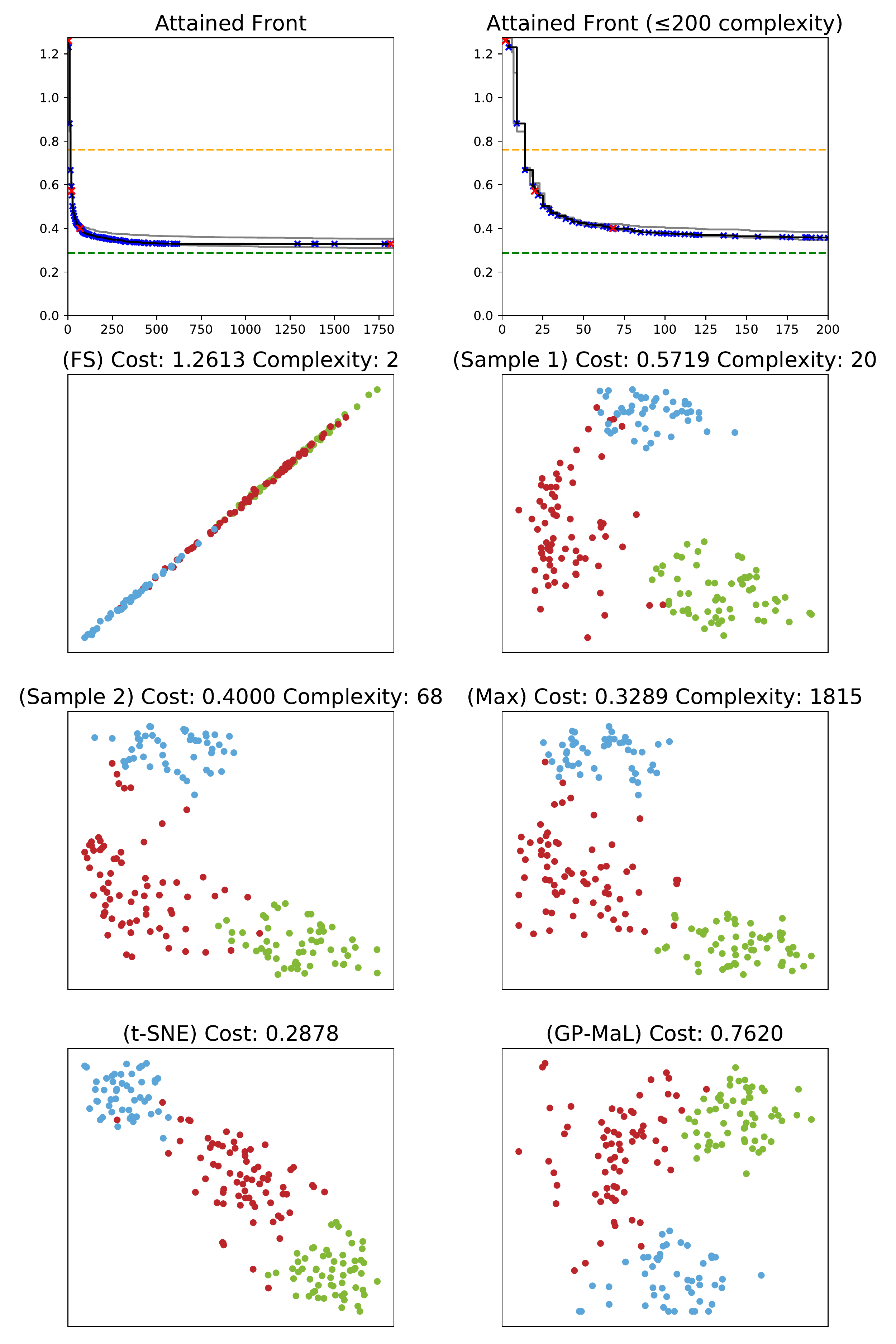}
		\put (-1,85.5) {(a)}
\put (31.5,85.5) {(b)}
\put (-1,60.5) {(c)}
\put (31.5,60.5) {(d)}
\put (-1,35.5) {(e)}
\put (31.5,35.5) {(f)}
\put (-1,10.5) {(g)}
\put (31.5,10.5) {(h)}
\end{overpic}
	\vspace{-1em}
	\caption{Wine.}
	\vspace{-1em}
	\label{fig:wine-median-PLOTS-PSO-PAPER}
\end{figure}

\Cref{fig:wine-median-PLOTS-PSO-PAPER} shows the results on another simple dataset: Wine. All three methods produce clear visualisations, with the main difference being that t-SNE projects all 3 classes \revisionOne{across the visualisation diagonally}, whereas the two GP methods produce a \revisionOne{curved pattern}. Even at a low level of complexity, GP-tSNE clearly separates the three classes, and higher levels of complexity tend to mostly refine the local structure within each class. As on the Iris dataset, the cost achieved by GP-tSNE at its maximum complexity is reasonably close to t-SNE, despite being produced by a mapping rather than just an embedding.

\begin{figure}[]
		\vspace{-1em}
	\centering
	\begin{overpic}[height=0.434\paperheight]{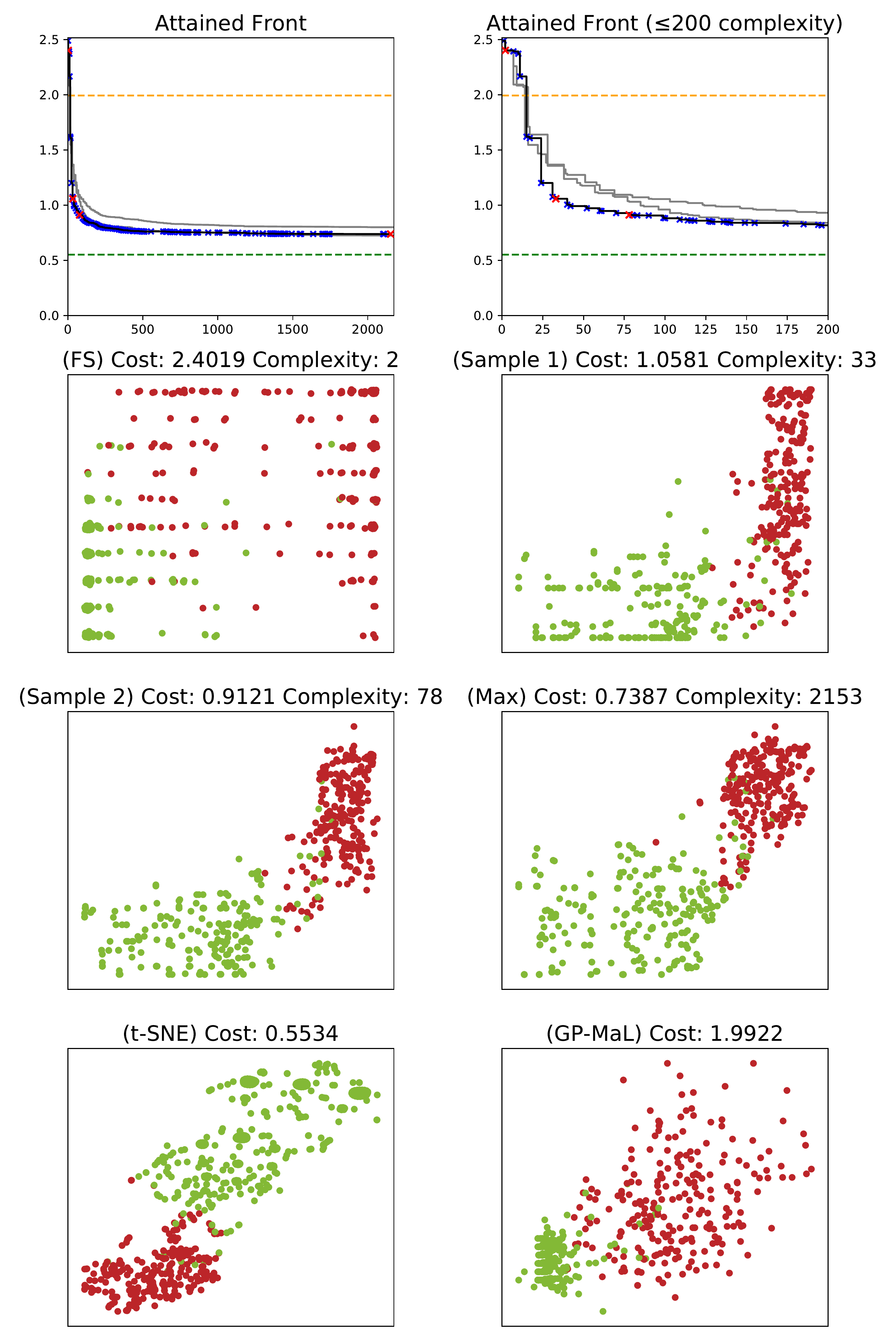}
		\put (-1,85.5) {(a)}
\put (31.5,85.5) {(b)}
\put (-1,60.5) {(c)}
\put (31.5,60.5) {(d)}
\put (-1,35.5) {(e)}
\put (31.5,35.5) {(f)}
\put (-1,10.5) {(g)}
\put (31.5,10.5) {(h)}	
		\end{overpic}
				\vspace{-1em}
	\caption{Breast Cancer Wisconsin.}
					\vspace{-1em}
	\label{fig:breast-cancer-wisconsin-median-PLOTS-PSO-PAPER}
\end{figure}
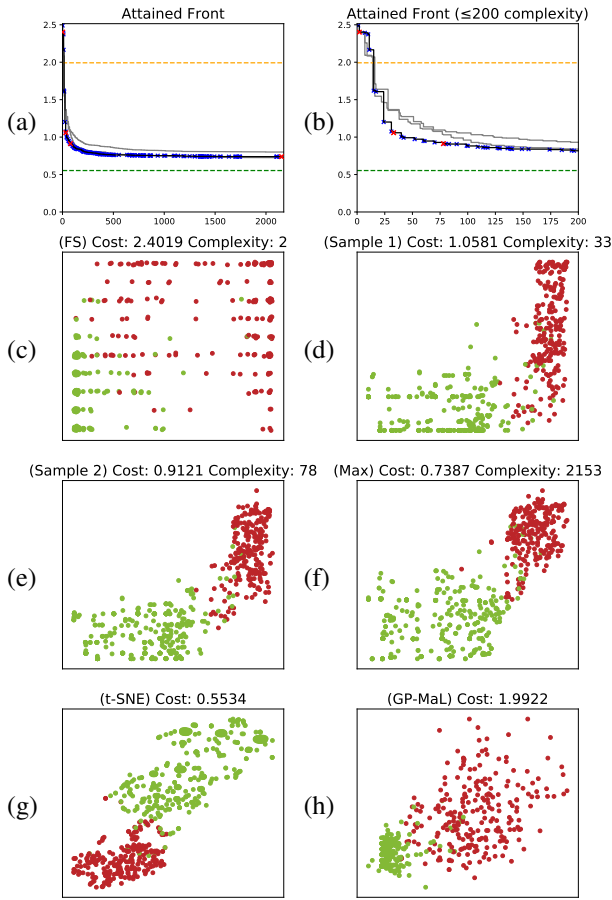

Both t-SNE and GP-tSNE clearly separate the two classes on the Wisconsin Breast Cancer dataset (\cref{fig:breast-cancer-wisconsin-median-PLOTS-PSO-PAPER}), whereas GP-MaL does not give as clear a separation boundary. GP-tSNE begins to show the division between the two classes as early as at a complexity of 33, with the red class becoming more tightly packed as the complexity increases further. Indeed, it appears that the GP models retain the same general ``approach'' in sample 1, 2 and at the maximum complexity, but become increasingly refined and accurate. In this way, it could be possible to use the simpler models to explore the more accurate patterns found by the bigger un-interpretable models by exploring how points move as complexity increases.

\begin{figure}[]
	\vspace{-1em}
	\centering
	\begin{overpic}[height=0.434\paperheight]{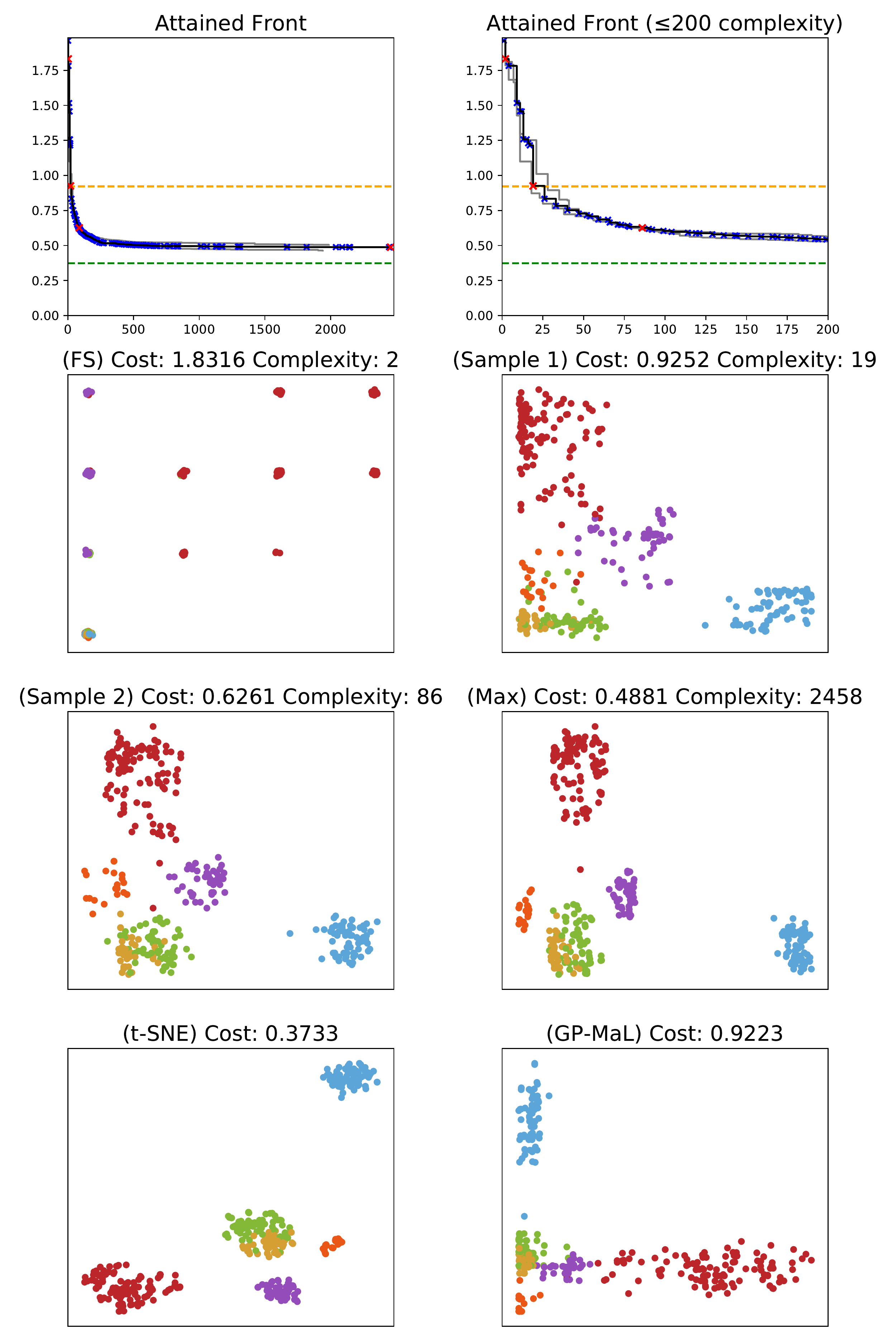}
		\put (-1,85.5) {(a)}
\put (31.5,85.5) {(b)}
\put (-1,60.5) {(c)}
\put (31.5,60.5) {(d)}
\put (-1,35.5) {(e)}
\put (31.5,35.5) {(f)}
\put (-1,10.5) {(g)}
\put (31.5,10.5) {(h)}	
	\end{overpic}
	\vspace{-1em}
	\caption{Dermatology.}
	\vspace{-1em}
	\label{fig:dermatology-median-PLOTS-PSO-PAPER}
\end{figure}
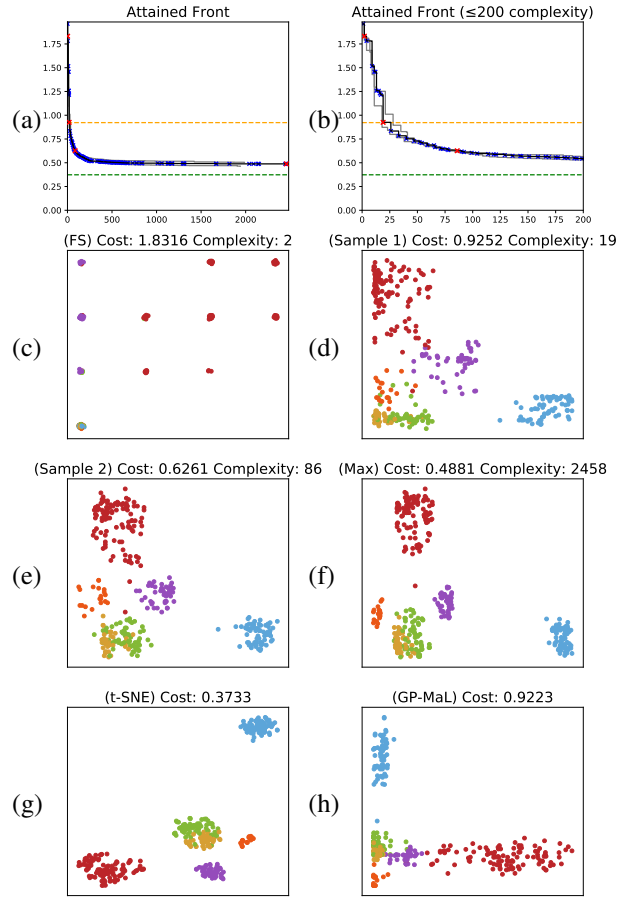

The Dermatology dataset (\cref{fig:dermatology-median-PLOTS-PSO-PAPER}) further reinforces this pattern: as GP-tSNE produces increasingly complex models, the visualisation becomes \revisionOne{of a }``higher-resolution'' and produces points that appear to be in a continuous space, rather than a discrete space. Sample 1 shows the blue class clearly being separated, and the red and purple classes beginning to separate from the other classes. Sample 2 shows a clear separation between all classes except the yellow and green ones, despite having a complexity of only 86. The maximum complexity separates the classes further, and compresses each class more tightly. The lowest cost for GP-tSNE gives a very similar visualisation to t-SNE, despite having a cost that is higher. GP-MaL is clearly worse than GP-tSNE, even when GP-tSNE has a tree size as low as 86.

%\Cref{fig:vehicle-median-PLOTS-PSO-PAPER} shows the visualisations for the Vehicle dataset, on which all three methods struggle to separate the four classes. This suggests that the class labels may not be representative of the underlying data. Some small patterns, such as the two groups of blue instances which are separated from the main set of instances, can be seen even at very low model complexity.
%\begin{figure}[]
%	\vspace{-1em}
%	\centering
%	\begin{overpic}[height=0.434\paperheight]{gpTSNEJoin/vehicle-median-PLOTS-PSO-PAPER}
%		\put (-1,85.5) {(a)}
%\put (31.5,85.5) {(b)}
%\put (-1,60.5) {(c)}
%\put (31.5,60.5) {(d)}
%\put (-1,35.5) {(e)}
%\put (31.5,35.5) {(f)}
%\put (-1,10.5) {(g)}
%\put (31.5,10.5) {(h)}		
%\end{overpic}
%	\vspace{-1em}
%	\caption{Vehicle.}
%	\label{fig:vehicle-median-PLOTS-PSO-PAPER}
%		\vspace{-1em}
%\end{figure}

\begin{figure}[]
	\vspace{-1em}
	\centering
	\begin{overpic}[height=0.434\paperheight]{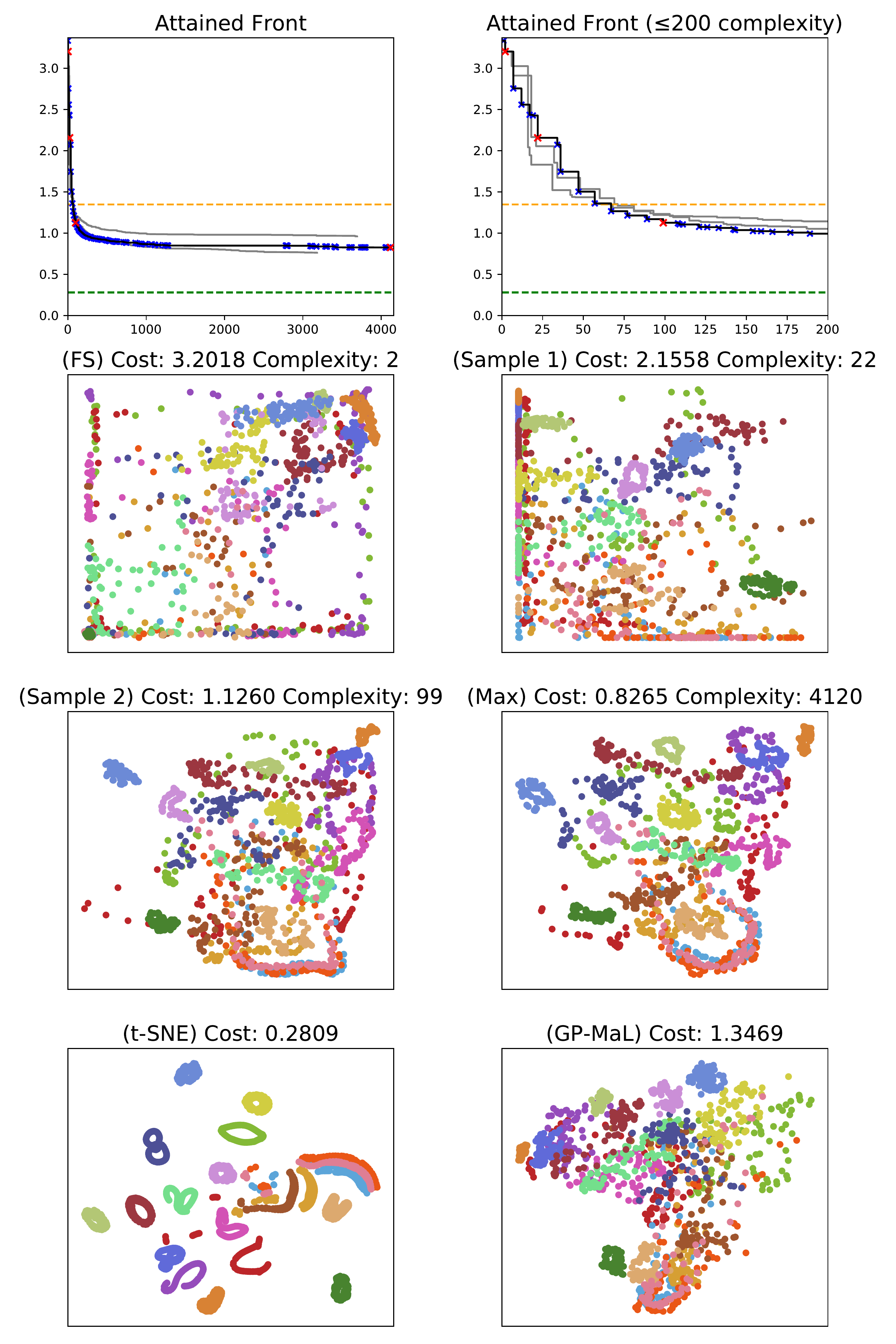}
		\put (-1,85.5) {(a)}
\put (31.5,85.5) {(b)}
\put (-1,60.5) {(c)}
\put (31.5,60.5) {(d)}
\put (-1,35.5) {(e)}
\put (31.5,35.5) {(f)}
\put (-1,10.5) {(g)}
\put (31.5,10.5) {(h)}	
\end{overpic}
	\vspace{-1em}
	\caption{COIL20.}
	\vspace{-1em}
	\label{fig:COIL20-median-PLOTS-PSO-PAPER}
\end{figure}

The COIL20 dataset (\cref{fig:COIL20-median-PLOTS-PSO-PAPER}) highlights t-SNE's ability to freely move points throughout the 2D space, as it clearly produces the best visualisation. It is interesting to note however that GP-tSNE is able to separate some classes well, even at low model complexity. For example, in Sample 1 (complexity of 22), a number of classes (dark green, pink, blue) are already clearly visible --- this suggests that these classes may be particularly well-defined, as they are easily separated. As the complexity increases, additional classes separate out, and the curve topology of the blue/pink/red classes seen in the t-SNE visualisation starts to form for GP-tSNE too. 

\begin{figure}[]
	\vspace{-1em}
	\centering
	\begin{overpic}[height=0.434\paperheight]{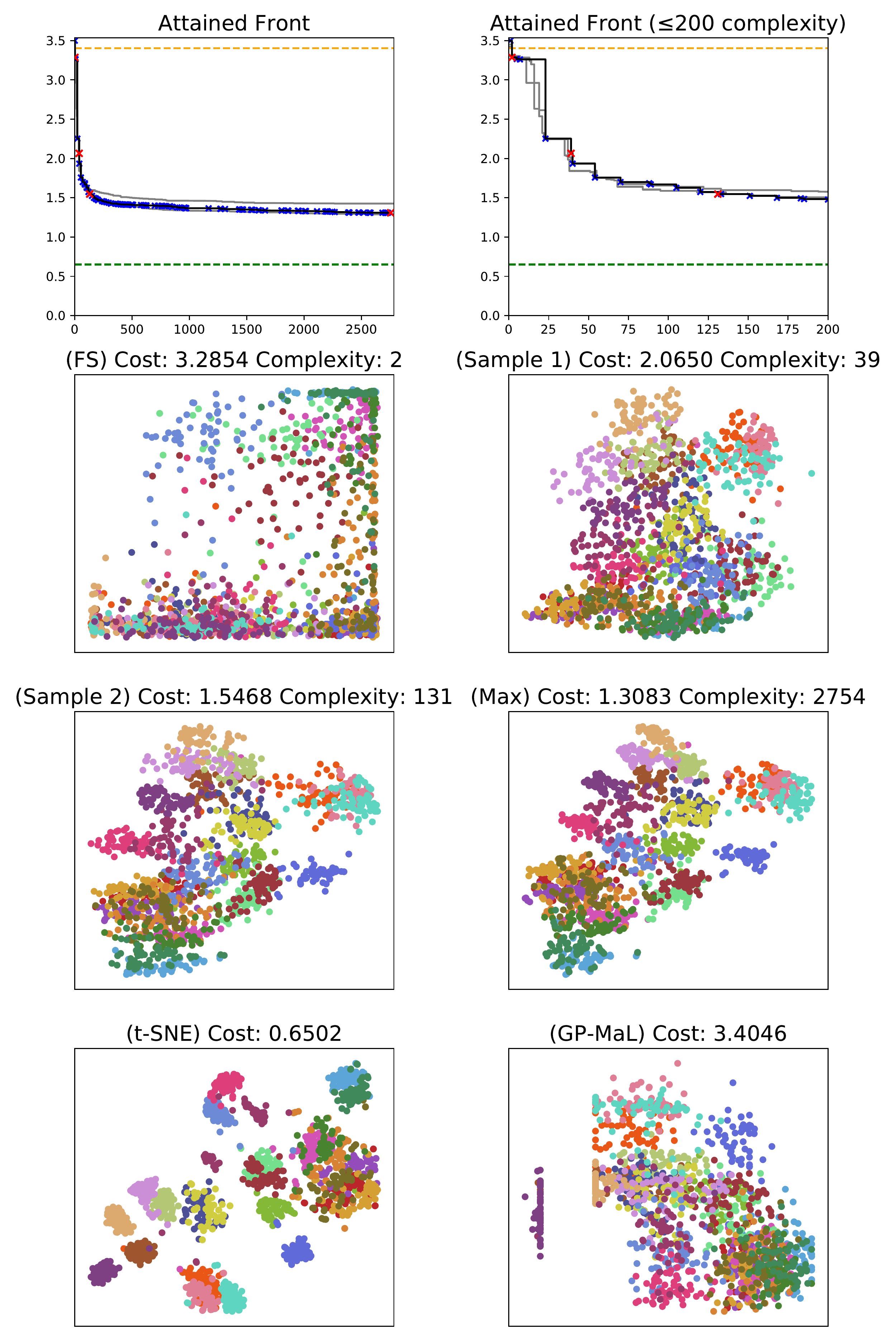}
		\put (-1,85.5) {(a)}
\put (31.5,85.5) {(b)}
\put (-1,60.5) {(c)}
\put (31.5,60.5) {(d)}
\put (-1,35.5) {(e)}
\put (31.5,35.5) {(f)}
\put (-1,10.5) {(g)}
\put (31.5,10.5) {(h)}	
\end{overpic}
	\vspace{-1em}
	\caption{Isolet.}
		\vspace{-1em}
	\label{fig:Isolet-median-PLOTS-PSO-PAPER}
\end{figure}

\Cref{fig:Isolet-median-PLOTS-PSO-PAPER} shows how the Isolet dataset has many overlapping classes, with only a few classes distinctly separated by t-SNE. The GP methods generally overlap the same classes as t-SNE, but do not manage to separate the groups quite as successfully. Isolet has the highest number of classes (26) of all the datasets, which makes it especially challenging for GP-based methods, since evolving two functions that can provide sufficient granularity to separate 26 classes is clearly very challenging and requires sufficiently complex trees. Despite this, it may be possible to gain some insight into why given classes overlap, as the general overlapping patterns start to be visible even at lower complexities of 39 to 131. 
\begin{figure}[]
	\vspace{-1em}
	\centering
	\begin{overpic}[height=0.434\paperheight]{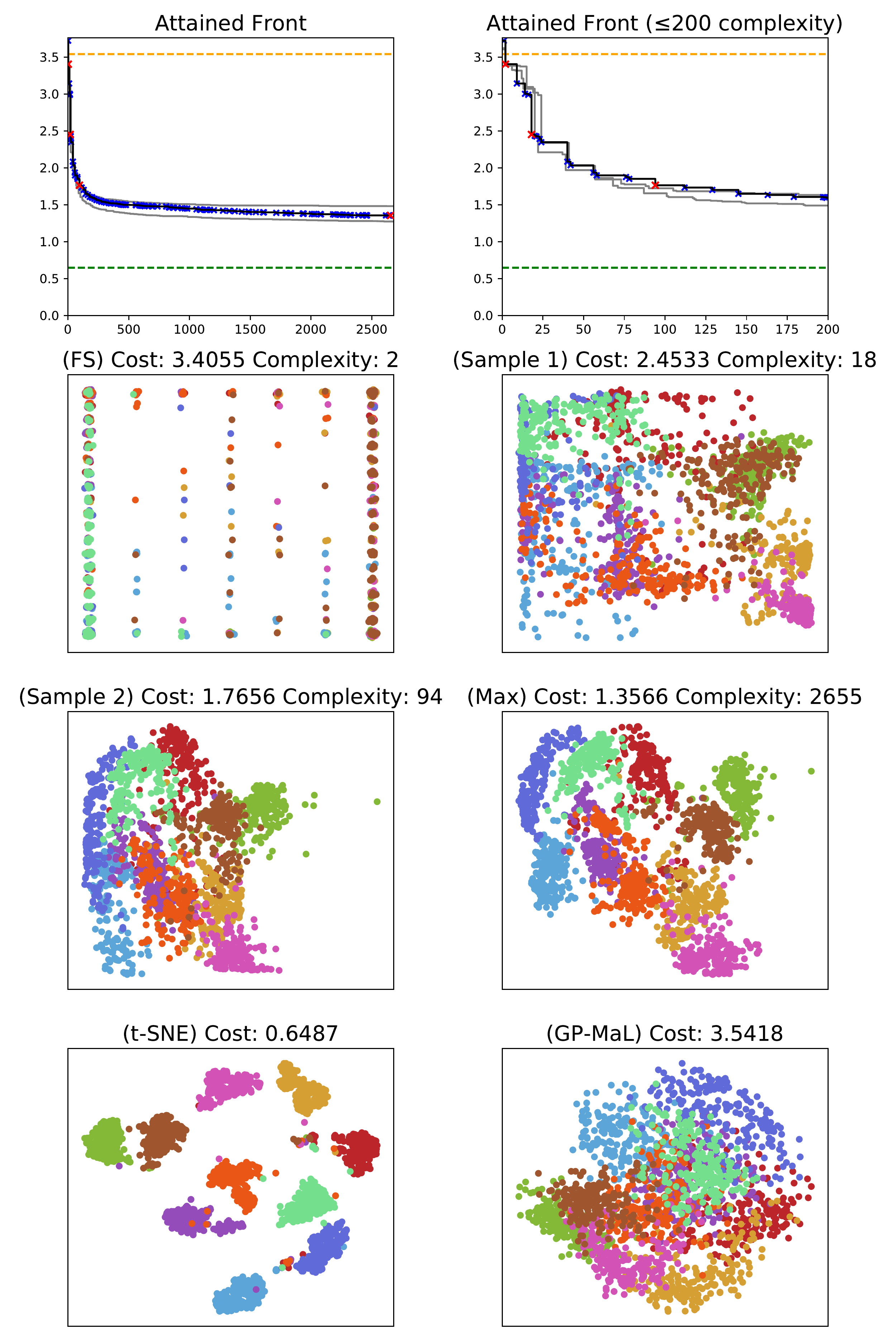}
		\put (-1,85.5) {(a)}
\put (31.5,85.5) {(b)}
\put (-1,60.5) {(c)}
\put (31.5,60.5) {(d)}
\put (-1,35.5) {(e)}
\put (31.5,35.5) {(f)}
\put (-1,10.5) {(g)}
\put (31.5,10.5) {(h)}	
\end{overpic}
	\vspace{-1em}
	\caption{MFAT.}
	\vspace{-1em}
	\label{fig:MFAT-median-PLOTS-PSO-PAPER}
\end{figure}
\begin{figure}[]
	\vspace{-1em}
	\centering
	\begin{overpic}[height=0.434\paperheight]{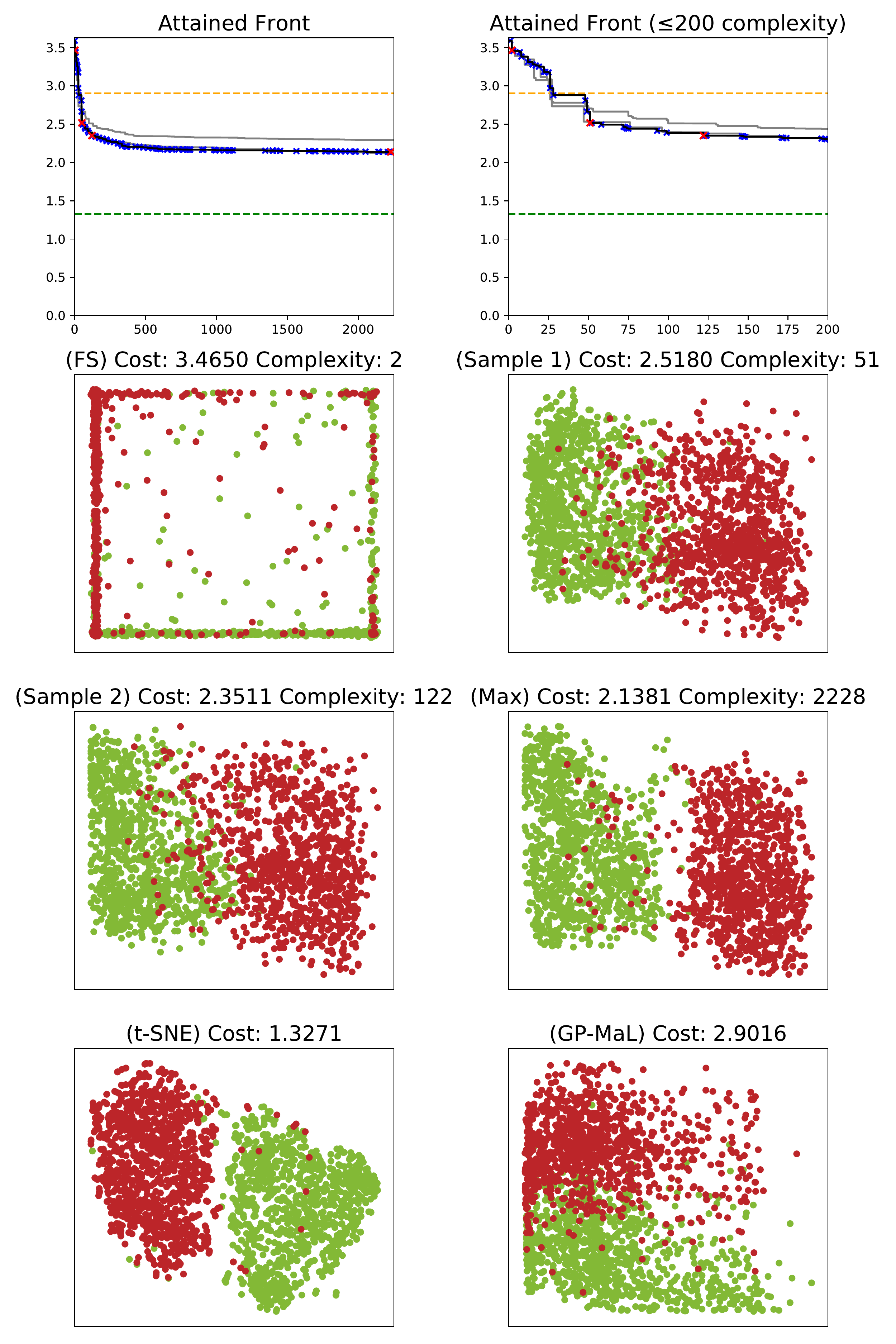}
		\put (-1,85.5) {(a)}
\put (31.5,85.5) {(b)}
\put (-1,60.5) {(c)}
\put (31.5,60.5) {(d)}
\put (-1,35.5) {(e)}
\put (31.5,35.5) {(f)}
\put (-1,10.5) {(g)}
\put (31.5,10.5) {(h)}	
\end{overpic}
	\vspace{-1em}
	\caption{MNIST 2-class.}
	\vspace{-1em}
	\label{fig:MNIST-2-median-PLOTS-PSO-PAPER}
\end{figure}
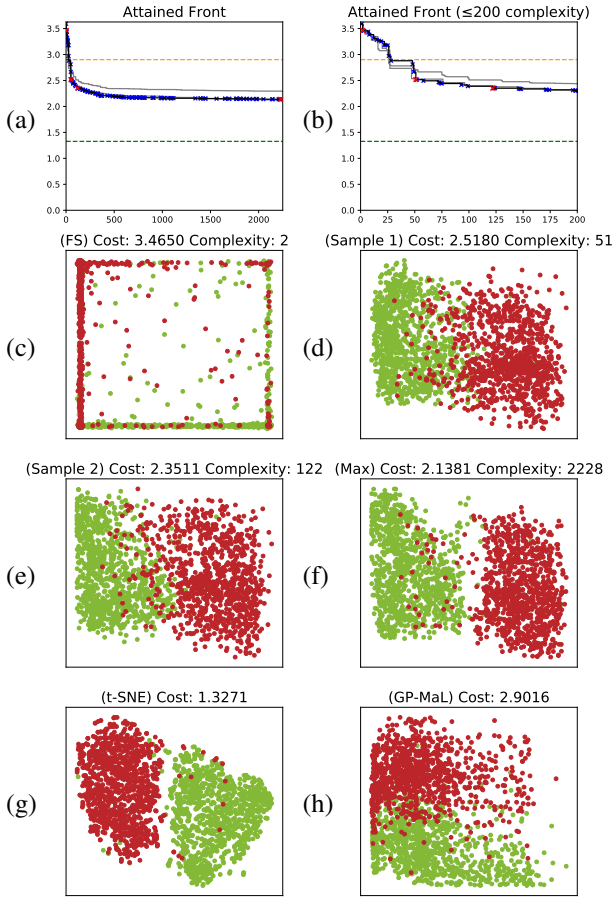

On the MFAT dataset (\cref{fig:MFAT-median-PLOTS-PSO-PAPER}), GP-tSNE is able to group each class effectively at the maximum complexity, but does not show the clear separation between classes that t-SNE provides. GP-tSNE clearly performs significantly better than GP-MaL, which overlaps the light green and purple classes onto other classes. The grouping of classes starts to appear at a complexity of 94, although the orange and purple classes overlap still. A complexity of 94 corresponds to two trees with 94 nodes in total --- which is clearly quite restrictive given that the MFAT dataset has 649 features and 10 classes. 

The visualisations in \cref{fig:MNIST-2-median-PLOTS-PSO-PAPER} (MNIST 2-class) provide an example of a large dataset with a small number of classes. All three methods are able to separate the two classes reasonably well, with GP-tSNE and t-SNE producing quite similar results. This separation is evident even at low model complexity in GP-tSNE -- at a complexity of 51, it is possible to draw a vertical line through the middle of the visualisation that would separate the two classes with reasonable accuracy. At higher levels of complexity, GP-tSNE is able to push the two classes apart to leave a clear gap between them. GP-MaL produces a less clear separation and struggles to distribute the points in the visualisation space well.

\begin{figure}[]
	\vspace{-1em}
	\centering
	\begin{overpic}[height=0.434\paperheight]{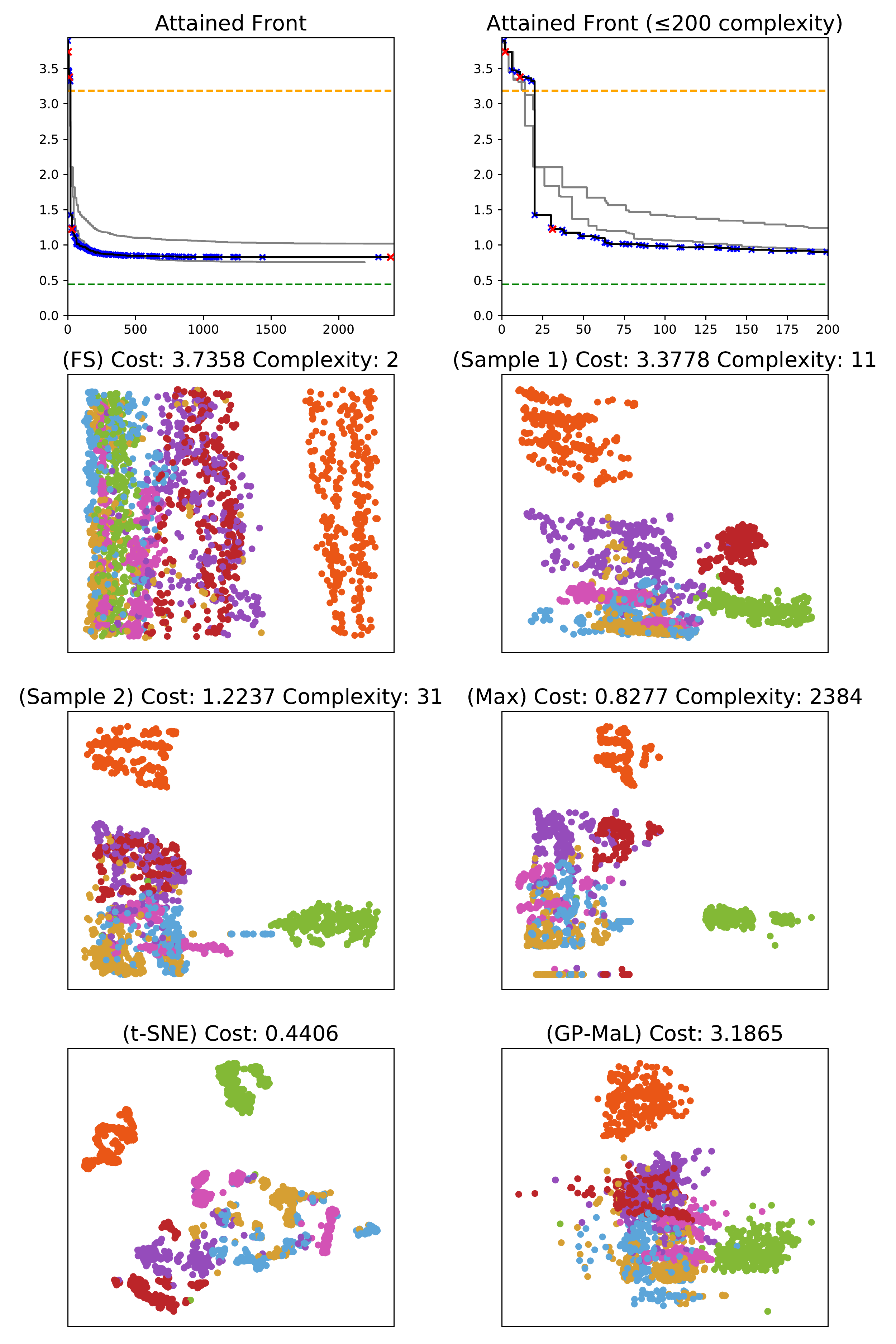}
		\put (-1,85.5) {(a)}
\put (31.5,85.5) {(b)}
\put (-1,60.5) {(c)}
\put (31.5,60.5) {(d)}
\put (-1,35.5) {(e)}
\put (31.5,35.5) {(f)}
\put (-1,10.5) {(g)}
\put (31.5,10.5) {(h)}	
\end{overpic}
	\vspace{-1em}
	\caption{Image Segmentation.}
	\vspace{-1em}
	\label{fig:image-segmentation-median-PLOTS-PSO-PAPER}
\end{figure}

On the final dataset, Image Segmentation (\cref{fig:image-segmentation-median-PLOTS-PSO-PAPER}), all three methods are able to separate the green and orange classes out, with t-SNE and GP-tSNE providing clearer separation margins. Both GP-tSNE and t-SNE also separate the red class to some extent, with t-SNE doing so slightly more effectively. An interesting observation is that the orange, and to a lesser extent the green classes, are separated at very low model complexity. The orange class is easily separated using only one feature, whereas the green class can be separated at a complexity of 31. This suggests that perhaps these are more natural/intrinsic classes in the dataset; perhaps they exhibit characteristics that make them particularly distinct from the other instances. Another interesting finding is that the red class is separated well by GP-tSNE at a complexity of 11 --- but then overlaps with the purple class at higher complexity. This may be a limitation of the t-SNE objective function or may suggest that the red class is actually a \textit{subclass} of the purple class and is incorrectly over-separated at a low complexity. \revisionOne{Indeed, the red class corresponds to pixels that are part of a path in an image, and the purple class to cement pixels. Clearly, many different things can be constructed out of concrete, including paths, which explains the overlap that is evident at higher complexities.} This phenomenon is another useful characteristic of GP-tSNE: it shows the different relationships between classes at different levels of model complexity.

\subsection{General Findings}

\begin{figure}[tb]
	\revisionOnePar{
		\centering
		\vspace{-.5em}
		\includegraphics[width=.8\linewidth]{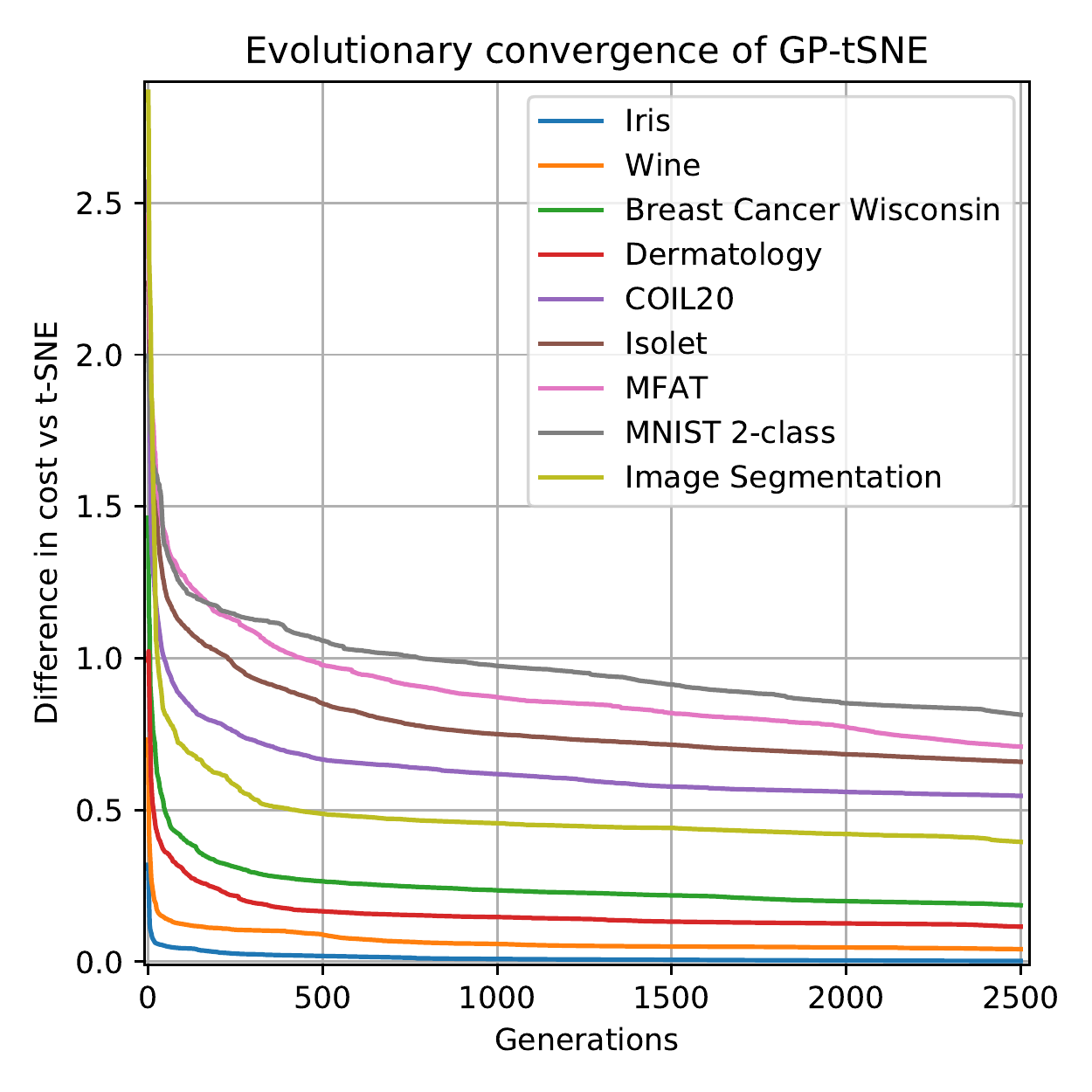}
		\vspace{-1em}
		\caption{Convergence of GP-tSNE over the evolutionary process: y-axis shows the difference in t-SNE cost between the best individual at a given generation and that attained by t-SNE.}
		\vspace{-1.5em}
		\label{convergenceAnalysis}	
	}
\end{figure}
The visualisations produced by GP-tSNE were consistently superior to those of our previous GP method, GP-MaL. GP-MaL was developed for more general MaL (i.e.\ not just reducing to two dimensions); by using a visualisation-specific quality measure, GP-tSNE was able to clearly improve. As the number of instances increased, t-SNE begun to produce clearer visualisations than GP-tSNE, although the same general patterns were shown by both methods. Given GP-tSNE is in essence tackling a strictly more difficult task, of evolving a \textbf{functional model} to produce a visualisation, we see this as a success for the first such approach to this task. \revisionOne{This pattern can be further seen in the convergence analysis shown in \Cref{convergenceAnalysis}. On the easier datasets (the first few in the legend), GP-tSNE converges before 1,000 generations. On the hardest datasets such as MNIST and MFAT, GP-tSNE is likely to benefit from additional generations}. 
\revisionOne{We are confident that future research in this new research direction will be able to further improve GP-tSNE on high-dimensional datasets with many classes.}

One particularly interesting observation was how GP-tSNE produced similar overall visualisations at different model complexities, with the more complex models being more granular/refined than the simpler ones. This demonstrates the advantages of an evolutionary approach: good sub-trees of a complex tree can be used to improve simpler trees (and vice-versa) via crossover, allowing for more efficient search that produces models with different levels of trade-off. In the following section, we will explore this phenomenon in more depth to highlight the novel advantages of GP-tSNE.
%\threebyside{13}{19}{26}

	\begin{figure*}[p!]
		\vspace{-2em}
	\centering
	\begin{minipage}[b]{.32\textwidth}
		\centering
		%\hspace{-.7cm}
		{\includegraphics[width=\textwidth]{{{gpTSNEJoin/dermatology13.0-0}}}\vspace{.5em}}
		\includegraphics[width=0.35\textwidth]{{{gpTSNEJoin/dermatology13.0-1}}}
		{\includegraphics[width=\textwidth]{{{gpTSNEJoin/dermatology13.0-PLOTS}}}	\vspace{-2.5em}}
	\end{minipage}
	\hfill %\hspace{-.7cm}
	\begin{minipage}[b]{.32\textwidth}
		{\includegraphics[width=\textwidth]{{{gpTSNEJoin/dermatology19.0-0}}}\vspace{.5em}}
		\includegraphics[width=\textwidth]{{{gpTSNEJoin/dermatology19.0-1}}}
		{\includegraphics[width=\textwidth]{{{gpTSNEJoin/dermatology19.0-PLOTS}}}	\vspace{-2.5em}}
	\end{minipage}
	\hfill %\hspace{-.7cm}
	\begin{minipage}[b]{.32\textwidth}
		{\includegraphics[width=\textwidth]{{{gpTSNEJoin/dermatology26.0-0}}}\vspace{-.15em}}
		\includegraphics[width=\textwidth]{{{gpTSNEJoin/dermatology26.0-1}}}
		{\includegraphics[width=\textwidth]{{{gpTSNEJoin/dermatology26.0-PLOTS}}}	\vspace{-2.5em}}
	\end{minipage}%
	
	\begin{minipage}[t]{.32\textwidth}
		\caption{Complexity of 13.}
		\label{fig:complexity13}
	\end{minipage}
	\hfill
	\begin{minipage}[t]{.32\textwidth}
		\caption{Complexity of 19.}
		\label{fig:complexity19}
	\end{minipage}
	\hfill
	\begin{minipage}[t]{.32\textwidth}
		\caption{Complexity of 26.}
		\label{fig:complexity26}
	\end{minipage}	
		\vspace{-1em}
\end{figure*}

\section{Further Analysis}

%\subsection{Interpretability of Solutions along the Front}
The results of GP-tSNE on the Dermatology dataset (\cref{fig:dermatology-median-PLOTS-PSO-PAPER}) were of particular interest as they showed similar results between GP-tSNE and t-SNE; showed the same general visualisation, but at different qualities across tree sizes; and achieved good visualisations even at reasonably small model complexities. To further demonstrate the value of GP-tSNE, this section analyses a selection of increasingly complex trees produced by the median GP run on this dataset.

\subsection{Simple Models}
%\twobyside{14}{20}

The simplest model which gives a reasonable visualisation is shown in \cref{fig:complexity13}, which contains one tree with three nodes, and the other with 10, for a total complexity of 13. Even with such a low complexity, three of the classes start to appear, with the blue, purple, and red classes already showing a clear separation from the others. The \revisionOne{top} tree, which gives the x-axis of the visualisation, can be written as $x = \text{f}_5+\text{f}_{11}+\text{f}_{32}+2\text{nf}_{14}-\text{f}_{21}-\text{nf}_{21}$. The \revisionOne{bottom} tree, which gives the y-axis is simply $y = \text{nf}_{20} - \text{f}_{15}$. These two trees use a total of seven unique features out of the 34 in the original feature set, yet they are able to separate three classes of the dataset well. In particular, the blue, red and purple classes can be separated out by drawing two vertical lines through the plot. In other words, two thresholds can be applied to the output of the \revisionOne{top} tree (x-axis) in order to roughly separate the classes into three groups: red; orange and green; and purple and blue. Given that all features have the same range of $[0,1]$ and the only feature subtracted in the tree is $\text{f}_{21}$, this suggests that the blue and purple classes have particularly small feature values for $\text{f}_{21}$ (as they have high x values), and the red class has particularly large values (as it has low x values). $\text{f}_{21}$ in the Dermatology dataset corresponds to the ``thinning of the suprapapillary epidermis'' feature, which is a feature commonly associated with the skin condition of psoriasis \cite{lever2015histopathology}. In the visualisation in \cref{fig:complexity13}, the red class corresponds to a diagnosis (class label) of psoriasis --- and indeed, the red instances appear on the left side of the plot, which is consistent with having a high value of $\text{f}_{21}$ being subtracted from the rest of the tree. This sort of analysis can be used by a clinician to understand that this feature alone could be used to diagnose psoriasis with reasonable accuracy, and also provides greater confidence in the visualisation's accuracy.

	\begin{figure*}[p!]
				%\vspace{-1em}
	%	\vspace{-1em}
	%\centering
	%\hspace{-.7cm}
	\begin{minipage}[b]{.32\textwidth}
		\flushleft
		{\includegraphics[width=.8\textwidth]{{{gpTSNEJoin/dermatology33.0-0}}}\vspace{.5em}}
		{\includegraphics[width=1.1\textwidth]{{{gpTSNEJoin/dermatology33.0-1}}}}
		{\centering \includegraphics[width=\textwidth]{{{gpTSNEJoin/dermatology33.0-PLOTS}}}	\vspace{-2.5em}}
	\end{minipage}
	\hfill %\hspace{0.3cm}
	\begin{minipage}[b]{.32\textwidth}
		\centering
		{\hspace*{-1cm}\includegraphics[width=1.45\textwidth]{{{gpTSNEJoin/dermatology59.0-0}}}\vspace{.5em}}
		{\hspace*{0.2cm} \includegraphics[width=1.2\textwidth]{{{gpTSNEJoin/dermatology59.0-1}}}}
		{\centering \includegraphics[width=\textwidth]{{{gpTSNEJoin/dermatology59.0-PLOTS}}}	\vspace{-2.5em}}
	\end{minipage}
	\hfill %\hspace{0.02cm}
	\begin{minipage}[b]{.32\textwidth}
		\flushright
		{\includegraphics[width=\textwidth]{{{gpTSNEJoin/dermatology104.0-0}}}\vspace{1.5em}}
		\includegraphics[width=0.82\textwidth]{{{gpTSNEJoin/dermatology104.0-1}}}
		{\centering \includegraphics[width=\textwidth]{{{gpTSNEJoin/dermatology104.0-PLOTS}}}	\vspace{-2.5em}}
\end{minipage} %\hspace{-.6cm} %
	
	\begin{minipage}[t]{.32\textwidth}
		\caption{Complexity of 33.}
		\label{fig:complexity33}
	\end{minipage}
	\hfill
	\begin{minipage}[t]{.32\textwidth}
		\caption{Complexity of 59.}
		\label{fig:complexity59}
	\end{minipage}
	\hfill
	\begin{minipage}[t]{.32\textwidth}
		\caption{Complexity of 104.}
		\label{fig:complexity104}
	\end{minipage}	
	\vspace{-1em}
\end{figure*}

When the model complexity is increased to 19 (\cref{fig:complexity19}), the separation of points within classes starts to become better-defined, and the orange class starts to become more distinct from the \revisionOne{yellow} and green classes.  The \revisionOne{top} tree actually uses fewer unique features (three) than at a complexity of 13, with significant weight put on $f_{32}$: $x = 4\text{nf}_{32} + \text{f}_{32} + 2\text{nf}_{14} + \text{nf}_{3} + \text{f}_{3}$, whereas the \revisionOne{bottom} tree uses four unique features: $y = \text{nf}_{20} + \text{f}_{20} + \text{nf}_{19} + \text{f}_{19} + \text{nf}_{8} - \text{nf}_{27}$, again for a total of seven unique features across both trees. The blue class (``lichen planus'') is clearly distinct from the other classes along the x-axis --- given that the \revisionOne{top} tree weights $\text{f}_{32}$ heavily, it seems likely that a high value of this feature is characteristic of the ``lichen planus'' diagnosis. Indeed, the dermatology literature commonly reports on this symptom being indicative of this diagnosis \cite{lever2015histopathology}.

The trees shown in \cref{fig:complexity26} appear very similar to the previous model, with the \revisionOne{top} tree varying only by the omission of one nf$_{14}$ node, and the introduction of the subtraction of the nf$_{8}$ feature. This is, however, sufficient to  cause some of the orange points to be separated from the yellow and green points along the x-axis --- indicating that higher values of nf$_8$ may indicate a point belongs to the orange class rather than the yellow one. The major change in the \revisionOne{bottom} tree is the introduction of the nf$_{21}$ and nf$_{6}$ features, which helps to compact the blue class and starts to separate the red class more strongly.

%represents an increase of only one complexity to the previous model, but with a very different appearance. The second tree is actually identical to the first tree of the previous model but has swapped position, causing an inversion of the visualisations from here on in. The first tree is surprisingly different to the second tree of the previous model, as it adds features $X21$ and $X30$ and removes the use of features $X15$. This change is sufficient to give a clearer separation between the red and purple classes, and between the purple and green/orange classes, which suggests that features $X21$ and $X30$ may be particularly characteristic of the purple class. 

\Cref{fig:complexity33} further pushes the blue class away from the other classes, at the expense of squashing the other classes together. This is achieved by making the \revisionOne{top} tree (x-axis) weight f$_{3}$ and f$_{32}$ even more strongly. It is interesting to note that all the trees analysed so far have been strictly linear functions, utilising only arithmetic and subtraction. This is perhaps not unexpected: utilising more sophisticated functions is likely to require complex trees to fully take advantage of them --- for example, taking the maximum of two simple sub-trees is unlikely to be more representative of the original dataset than simply adding together features. This also has the benefit of making the simpler trees easier to interpret and understand. While there are many methods for doing linear transformations for visualisation, such as PCA, these methods generally function by weighting \textbf{all} features to different extents; GP-tSNE only selects a subset of features to be used, and so is inherently more interpretable. The sequential analysis of increasingly more complex models clearly provides additional insight that would be lacking in a non-multi-objective approach. This analysis technique is also an exclusive benefit of GP-tSNE among other visualisation techniques which are primarily black boxes with one visualisation produced per run.

%further increases the amount of detail in the visualisation. The second model (y-axis) is still semantically unchanged from the first model of \cref{fig:complexity20}. The first model (x-axis) however, now uses a total of seven unique features: $X6,16,19,20,21,27,29$. This allows it to spread the blue class out further, as well as adding clear distinction between the green and orange classes which was not present before. The yellow class also begins to separate somewhat.

\subsection{Complex Models}

%\twobyside{60}{122}

From here, the improvement in quality with the addition of more complexity begins to show clear diminishing returns, with the cost decreasing by only about 0.1 as the complexity is roughly doubled from 33 to 59, and from 59 to 104. \Cref{fig:complexity59} gives a visualisation that is very similar to that of the maximum complexity (see \cref{fig:dermatology-median-PLOTS-PSO-PAPER}), with the yellow and green classes overlapping but all other classes clearly distinctly separated. This is also the first stage at which constants and non-linear operators such as $-$ and $\max$ are introduced. The trees used in previous models are still recognisable as sub-trees (or ``building blocks'') in the more complex trees in \cref{fig:complexity59} --- it may be possible to analyse the more complex trees based on what has been added in addition to these sub-trees. \Cref{fig:complexity104} does not change the class separations present in \cref{fig:complexity59}, but better separates within-class instances, as well as moving the purple class away from the other classes. 

We do not attempt to analyse models with a complexity of over 100 in detail, as the trees become difficult to interpret easily, and eventually become as much of a black box as t-SNE itself is. The main aim of this paper is to produce good-quality (not state-of-the-art) visualisations that are \textit{interpretable} in order to provide deeper insight into the relationships within a dataset in terms of the features it uses. If visualisation quality is the primary goal, then methods such as t-SNE are expected to be more appropriate, given they are not constrained to finding a functional mapping from the original feature space to the visualisation axes. While the very complex GP trees cannot be feasibly interpreted, they are still important to the GP-tSNE algorithm. During the evolutionary process, it is common to see complex trees with low cost being automatically simplified through the removal of superfluous sub-trees via crossover and mutation, allowing reductions in cost to ``trickle down'' to simpler models along the front. When we restricted the EMO search to only simple trees, we found that results were much poorer, with smaller trees having much higher cost. The complex trees are also useful outright as they can be directly applied for visualising future instances (e.g.\ streaming data), whereas t-SNE must re-optimise its embedding each time.

\subsection{Summary}
In this section, we showed how progressive examination of the \revisionOne{approximation} front from least to most complex models allowed insight into the structure of the data that is not easily achievable through traditional visualisation algorithms. Using the Dermatology dataset as a case study, we showed that even simple models could separate out some classes clearly, with the use of only a few features. As we increased the complexity, we found that the granularity/local structure within classes improved, but the overall patterns changed only slowly. In this way, it is possible to use simple, understandable models to provide insight into how more complex models are able to produce good-quality visualisations.
\vspace{-.25em}
\section{Conclusion}
This paper highlighted the need for research into machine learning methods that can not only produce clear visualisations, but do so through a model that can be interpreted itself to give insight into how the visualisation arose from the features of the dataset. The use of GP was suggested due to its functional model structure, and a multi-objective approach was proposed that gave a staggered front of visualisations with different levels of trade-off between visual clarity and model complexity. Results on nine different datasets highlighted the promise in using a GP approach for this task, with visualisations produced that showed clear characteristics of datasets even at model complexities that could be low enough to understand. This was further showcased through an in-depth analysis of an evolved front on the Dermatology dataset, where concrete insight into how the dataset was structured was found by examining a range of models with different complexities.

As this is the first multi-objective GP approach to this problem, there is a clear need for continued future research. While the quality of visualisations were encouraging, on more complex datasets they fell short of those produced by t-SNE. This is not unexpected given t-SNE does not produce a functional mapping, but rather an embedding of the data. The measure of visualisation quality used in this work was based on the one used by t-SNE given it is the state-of-the-art measure. However, the design of t-SNE was constrained by the need for a differentiable cost function; as an EC-based method, GP-tSNE need not have this constraint on its objective function --- there is likely to be better measures of visualisation quality that could be used in an EC context. It was also found that the more unique features used by a GP tree, the more difficult it was to understand; some sort of evolutionary pressure towards trees using a small set of (cohesive) features may improve this.

\vspace{-.25em}

\bibliographystyle{IEEEtran}
\bibliography{biblo}
%

\vspace{-3.5em}
\begin{IEEEbiography}[{\includegraphics[width=1in,height=1.25in,clip,keepaspectratio]{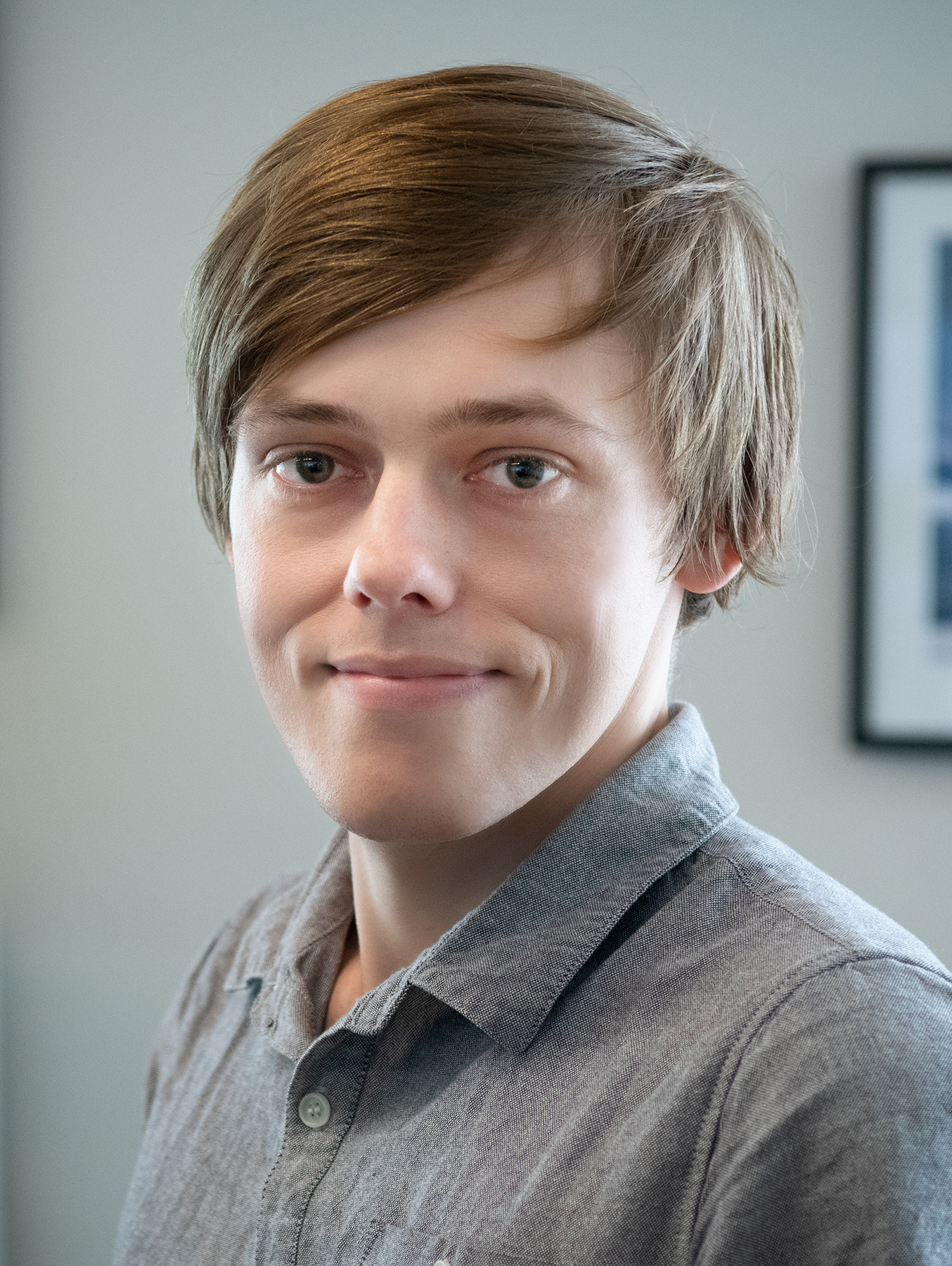}}]{Andrew Lensen} (M'17) received the B.Sc., B.Sc.(Hons 1$^{\text{st}}$ class), and Ph.D. degrees in computer science from Victoria University of Wellington, Wellington, New Zealand, in 2015, 2016, and 2019, respectively. 
	
He is currently a Post-doctoral Research Fellow in the Evolutionary Computation Research Group within the School of Engineering and Computer Science at Victoria University of Wellington. His current research interests are mainly in the use of evolutionary computation for feature manipulation in unsupervised learning, with a particular focus on the use of genetic programming for manifold learning, clustering, and feature synthesis. 

Dr Lensen serves as a regular reviewer of several international conferences, including IEEE Congress on Evolutionary Computation, and international journals such as the IEEE Transactions on Evolutionary Computation, and the IEEE Transactions on Cybernetics.
\end{IEEEbiography}
\vspace{-3.5em}
\begin{IEEEbiography}[{\includegraphics[width=1in,height=1.25in,clip,keepaspectratio]{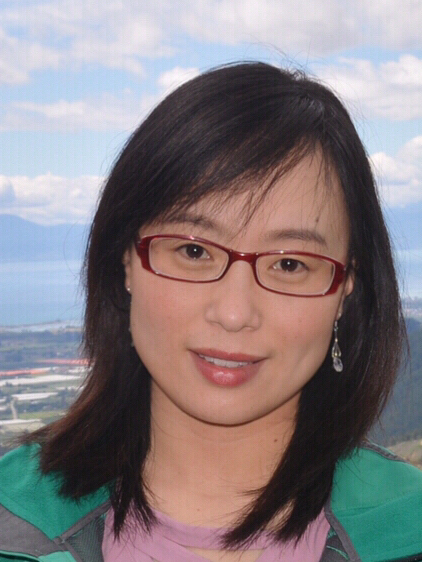}}]{Bing Xue} (M’10) received the B.Sc.\ degree from the Henan University of
	Economics and Law, Zhengzhou, China, in 2007, the M.Sc.\ degree in
	management from Shenzhen University, Shenzhen, China, in 2010, and the
	PhD degree in computer science in 2014 at Victoria University of
	Wellington, New Zealand. 
	
	She is currently an Associate Professor in
	the School of Engineering and Computer Science at Victoria University of
	Wellington. She has over 200 papers published in fully refereed
	international journals and conferences  and her research focuses mainly
	on evolutionary computation, feature selection, feature
	construction,  image analysis, and transfer learning.
	
	Dr Xue is
	currently the Chair of IEEE Computational Intelligence Society
	(CIS) Data Mining and Big Data Analytics Technical Committee, Chair
	of the IEEE Task Force on Evolutionary Feature Selection and
	Construction, and Vice-Chair of IEEE CIS Task Force on Transfer Learning
	\& Transfer Optimization,  and  Vice-Chair of IEEE CIS Task Force  on
	Evolutionary Deep Learning and Applications. 
\end{IEEEbiography}
\vspace{-3.5em}
\begin{IEEEbiography}[{\includegraphics[width=1in,height=1.25in,clip,keepaspectratio]{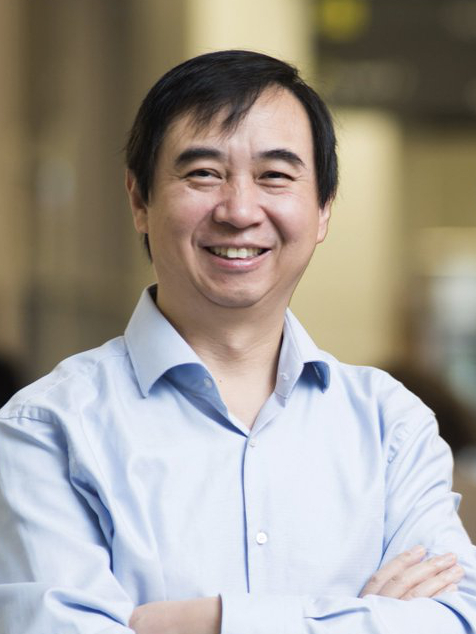}}]{Mengjie Zhang} (M’04–SM’10-F'19) received the B.E.\ and M.E.\ degrees
	from the Artificial Intelligence Research Center, Agricultural University
	of Hebei, Hebei, China, and the Ph.D.\ degree in computer science from
	RMIT University, Melbourne, Australia, in 1989, 1992, and 2000,
	respectively. 
	
	He is currently Professor of Computer Science, Head of the
	Evolutionary Computation Research Group, and the Associate Dean
	(Research and Innovation) in the Faculty of Engineering. His current
	research interests include evolutionary computation, particularly
	genetic programming, particle swarm optimization, and learning
	classifier systems with application areas of image analysis,
	multi-objective optimization, feature selection and reduction, job shop
	scheduling, and transfer learning. He has published over 500 research
	papers in refereed international journals and conferences. 
	
	Prof.\ Zhang
	is a Fellow of the Royal Society of New Zealand and have been a Panel member
	of the Marsden Fund (New Zealand Government Funding), a Fellow of IEEE,
	and a member of ACM. He was the chair of the IEEE CIS Intelligent
	Systems and Applications Technical Committee, and chair for the IEEE CIS
	Emergent Technologies Technical Committee and the Evolutionary
	Computation Technical Committee, and a member of the IEEE CIS Award
	Committee. He is a vice-chair of the IEEE CIS Task Force on Evolutionary
	Feature Selection and Construction, a vice-chair of the Task Force on
	Evolutionary Computer Vision and Image Processing, and the founding
	chair of the IEEE Computational Intelligence Chapter in New Zealand. He
	is also a committee member of the IEEE NZ Central Section.
\end{IEEEbiography}
\vspace{-.5em}

%\newpage



%\vfill


%\appendix
%\newpage
%\includepdf[pages=-,scale=.95]{/home/lensenandr/gpTSNETake3/subset/resultSamples.pdf}

\end{document}